\documentclass[conference]{IEEEtran}
\IEEEoverridecommandlockouts
\usepackage{cite}
\usepackage{amsmath,amssymb,amsfonts}
\usepackage{algorithmic}
\usepackage{graphicx}
\usepackage{diagbox}
\usepackage{textcomp}
\usepackage{xcolor}
\def\BibTeX{{\rm B\kern-.05em{\sc i\kern-.025em b}\kern-.08em
		T\kern-.1667em\lower.7ex\hbox{E}\kern-.125emX}}
	
\usepackage{caption}
\usepackage{subcaption}
\usepackage[ruled,vlined]{algorithm2e}
\usepackage{multirow}

\DeclareMathOperator*{\argmin}{arg\,min}

\begin{document}

\title{BinarizedAttack: Structural Poisoning Attacks to Graph-based Anomaly Detection}

\author{\IEEEauthorblockN{Yulin Zhu\IEEEauthorrefmark{1}, Yuni Lai\IEEEauthorrefmark{1}, Kaifa Zhao\IEEEauthorrefmark{1}, Xiapu Luo\IEEEauthorrefmark{1}, Mingquan Yuan\IEEEauthorrefmark{2}, Jian Ren\IEEEauthorrefmark{3}, Kai Zhou\IEEEauthorrefmark{1}}
\IEEEauthorblockA{\IEEEauthorrefmark{1}\textit{Dept. of Computing}, \textit{The Hong Kong Polytechnic University}, HKSAR\\ 
yulinzhu@polyu.edu.hk, lenalotus97@gmail.com, cskzhao@comp.polyu.edu.hk,\\
csxluo@comp.polyu.edu.hk, kaizhou@comp.polyu.edu.hk}
\IEEEauthorblockA{\IEEEauthorrefmark{2}\textit{Sam's Club Innovation Center}, Dallas, TX, USA, mingquan.yuan@samsclub.com }
\IEEEauthorblockA{\IEEEauthorrefmark{3}\textit{Dept. of Electrical and Computer Engineering}, \textit{Michigan State University}, East Lansing, MI, USA, renjian@msu.edu}
}

\maketitle

\begin{abstract}
	Graph-based Anomaly Detection (GAD) is becoming prevalent due to the powerful representation abilities of graphs as well as recent advances in graph mining techniques. These GAD tools, however, expose a new attacking surface, ironically due to their unique advantage of being able to exploit the relations among data. That is, attackers now can manipulate those relations (i.e., the structure of the graph) to allow some target nodes to evade detection.
    In this paper, we exploit this vulnerability by designing a new type of targeted structural poisoning attacks to a representative regression-based GAD system termed $\mathsf{OddBall}$. Specifically, we formulate the attack against $\mathsf{OddBall}$ as a bi-level optimization problem, where the key technical challenge is to efficiently solve the problem in a discrete domain.  We propose a novel attack method termed $\mathsf{BinarizedAttack}$ based on gradient descent. Comparing to prior arts, $\mathsf{BinarizedAttack}$ can better use the gradient information, making it particularly suitable for solving combinatorial optimization problems. Furthermore, we investigate the attack transferability of $\mathsf{BinarizedAttack}$ by employing it to attack other representation-learning-based GAD systems. Our comprehensive experiments demonstrate that $\mathsf{BinarizedAttack}$ is very effective in enabling target nodes to evade graph-based anomaly detection tools with limited attacker's budget, and in the black-box transfer attack setting, $\mathsf{BinarizedAttack}$ is also tested effective and in particular, can significantly change the node embeddings learned by the GAD systems. Our research thus opens the door to studying a new type of attack against security analytic tools that rely on graph data.

\end{abstract}



\section{Introduction}
Anomaly detection is a long-standing task in the field of data science and engineering with the goal to spot unusual patterns from the massive amount of data. Recently, due to the powerful representation abilities of graphs as well as the advances in graph mining and learning techniques, Graph-based Anomaly Detection (GAD) is becoming prevalent across a wide spectrum of domains. Various GAD systems are proposed and deployed as an indispensable security component in detecting, for example, fake accounts in social networks \cite{8621913}, fraudulent transactions in the financial industry \cite{fraudpayments}, and Botnets in communication networks \cite{180611}. 

Those GAD systems, however, expose a new attacking surface, ironically due to their unique advantage of being able to exploit the connections among data (i.e., edges in graphs). For example, in social network analysis, an attacker can proactively manage her social ties, by adding or deleting friendship connections, to reduce the probability of being detected as malicious. That is, adversaries now can \textit{evade} GAD via manipulating the graph structure; we term this class of attacks as \textit{structural attacks}.

\begin{figure}[t]
	\centering
	\includegraphics[scale=0.45]{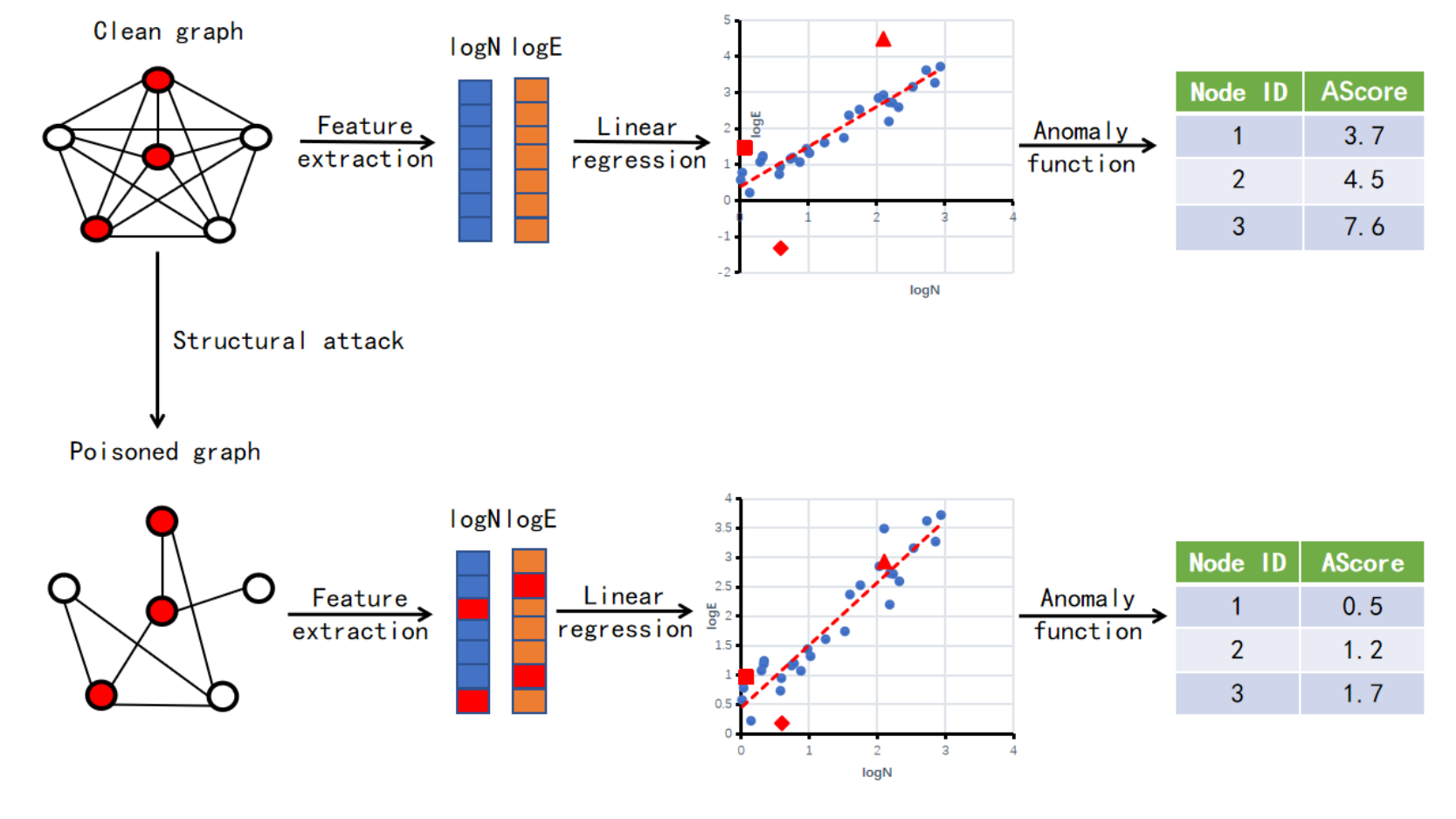}
	\caption{Illustration of the structural attack to $\mathsf{OddBall}$: $\mathsf{OddBall}$ extracts features from the graph structure, using regression-based approaches to compute the anomaly scores of nodes; structural attacks then poison the graph structure, equivalently manipulating the extracted features and finally resulting in misleading anomaly scores. This example shows that structural attacks significantly decrease the anomaly scores of some target nodes (red), allowing them to possibly evade the detection of $\mathsf{OddBall}$.}
	\label{fig-attack-procedure}
\end{figure}

We thus initiate the study of structural attacks on GAD with the goal of answering to what extend an attacker can evade the detection of prevalent GAD tools by solely manipulating graph structures. Studying such structural attacks has significant practical implications. Previous research on evading detection tools (e.g., PDF/Android malware detection \cite{7791883}) primarily operate in a feature space, where attackers manipulate the \textit{extracted} feature vectors. Consequently, it is known that the manipulated feature vectors are hard to be mapped back to real entities (e.g., PDF/Android software), affecting the effectiveness of the actual attacks. 
In contrast, altering the structure of a graph always corresponds to manipulating actual connections among entities (e.g., friendships in social networks), making structural attacks highly realizable in practice. Besides, there are scenarios where an attacker could have control of the whole graph structure,  allowing the global optimization of the attacks. A representative example is the Command \& Control center in Botnets \cite{180611}, which can coordinate the communication among Bots, i.e., globally optimizing the structure of the communication graph, to evade Botnets detection tools. These practical threats imposed by structural attacks motivate us to systematically study the attacker's ability to evade GAD.


In this paper, we instantiate our study by attacking a family of regression-based GAD algorithms termed $\mathsf{OddBall}$~\cite{oddball}. In a nutshell (illustrated in Fig.~\ref{fig-attack-procedure}), $\mathsf{OddBall}$ first extracts \textit{structural features} of each node in the graph, then builds regression models over those features, and finally computes the node anomaly scores. We focus on $\mathsf{OddBall}$ mainly due to the following reasons. First, $\mathsf{OddBall}$ stands for a class of simple while effective approaches that require much less information from data, thus are widely adopted in practice. Second, the hand-crafted features in $\mathsf{OddBall}$  are the results of domain-specific knowledge, which well captures the intrinsic structural characterizes of anomalies in the graph. In fact, the features extracted in $\mathsf{OddBall}$  are also used as input in other unsupervised deep-learning-based methods. In light of this, attacking $\mathsf{OddBall}$ can also serve as a good proxy for attacking other GAD tools. Indeed, we conducted such transfer attacks in Section~\ref{sec-transfer}. Fig~\ref{fig-attack-procedure} illustrates the main idea of attacking $\mathsf{OddBall}$. By modifying the graph structure, an attacker is equivalently changing the node features extracted by $\mathsf{OddBall}$. Consequently, the regression model built upon those features would falsely output node anomaly scores.

However, attacking GAD systems such as $\mathsf{OddBall}$ faces several challenges. First, GAD systems naturally feature an unsupervised/semi-supervised setting, where all nodes reside in a single graph. Attacking GAD thus is poisoning the data from which the detection model is built, which is in strong contrast to those attacks that are manipulating the inputs to a \textit{fixed trained} model. These \textit{poisoning attacks} are mathematically modeled as bi-level optimization problems, which are notoriously hard to solve in the literature.  Second, structural attacks operate in a discrete domain, resulting in a strategy space exponential in the graph size. Furthermore, the discrete nature makes it hard to adopt gradient-descent-based attack methods \cite{Z_gner_2018,DBLP:journals/corr/abs-1902-08412}, as the attacker now needs to make binary decisions on deleting or keeping the connections in the graph. Indeed as we will show later, trivially utilizing the gradient information may result in non-effective attacks. 

To tackle these challenges, we propose $\mathsf{BinarizedAttack}$, which is  inspired by \textit{Binarized Neural Networks} (BNN)~\cite{courbariaux2016binarized} designed for model compression. Specifically, BNN transforms the real-valued model weights to discrete values $+1$/$-1$ to reduce the model size. To find the discrete optimal weights, BNN uses a continuous version of the weights when computing the gradients in the backward propagation. In light of this, $\mathsf{BinarizedAttack}$ associates a \textit{discrete} as well as a \textit{continuous} decision variable for each edge/non-edge in the graph and uses a novel gradient-descent-based approach to solve the resulting discrete optimization problem. Specifically, in the forward pass, the discrete decision variables are used to evaluate the objective function. In the backward pass, the continuous decision variables are firstly updated based on the \textit{fractional} gradients, and the discrete ones are then updated accordingly. In essence, $\mathsf{BinarizedAttack}$ could better use the gradient information to guide the search for the optimal graph structure. Comprehensive experiments demonstrate that $\mathsf{BinarizedAttack}$ can effectively solve the discrete optimization problem and outperforms other gradient-descent-based approaches.


We further study the \textit{attack transferability} of $\mathsf{BinarizedAttack}$ to other GAD systems. In practice, the actual deployed GAD system could possibly be sensitive information that could not be accessible to attackers. Thus, it is of practical interests and significance to know how an attack method purposely designed for a target system would perform against other systems. In this paper, we use $\mathsf{BinarizedAttack}$ designed for $\mathsf{OddBall}$ to attack two other GAD systems ({GAL \cite{10.1145/3340531.3411979} and ReFeX \cite{ReFeX}}) in a \textit{black-box} manner.
Our intuition is that as $\mathsf{OddBall}$ can well capture the structural patterns that are generically important for identifying anomalies, $\mathsf{BinarizedAttack}$ has the ability to obfuscate those patterns thus possibly evading any GAD systems that rely on graph structures. Comprehensive experiments in Section~\ref{sec-transfer-GAL} demonstrate good attack transferability of $\mathsf{BinarizedAttack}$.

Finally, we summarize the main contributions of this paper as follows:
\begin{itemize}
	\item \textbf{New Vulnerability}. We identify and initiate the study of a new vulnerability of graph-based anomaly detection systems. Our results show that by slightly modifying the graph structure, attackers can successfully evade GAD systems.  The research opens the door to studying this new type of attacks against graph-based learning systems.
	\item \textbf{Effective Structural Poisoning Attacks}. We model structural poisoning attacks as bi-level discrete optimization problems. Technically, we propose a novel gradient-descent-based approach $\mathsf{BinarizedAttack}$ that can effectively solve discrete optimization problems, outperforming existing methods.
	\item \textbf{Attack Transferability}. We conduct comprehensive experiments demonstrating that $\mathsf{BinarizedAttack}$ is also effective in attacking  other GAD systems in a \textit{black-box} setting, making $\mathsf{BinarizedAttack}$ highly practical. 	
\end{itemize}
The rest of the paper is organized as follows. The related works are highlighted in Section~\ref{sec-related}. We introduce the target GAD system in Section~\ref{sec-GAD} and formulate the poisoning structural attacks in Section~\ref{sec-formulation}. The attack methods are introduced in Section~\ref{sec-methods}. We further investigate the attack transferability of the proposed method and explore possible countermeasures in Section~\ref{sec-transfer} and Section~\ref{sec-defense}, respectively. We conduct comprehensive experiments to evaluate our methods from several aspects in Section~\ref{sec-exp}. Finally, we conclude in Section~\ref{sec-conclude}.

\section{Related Works}
\label{sec-related}
\textbf{Graph-based anomaly detection}. The essential task of Graph-based Anomaly Detection (GAD) is to identify anomalous patterns from graph data. It becomes prevalent due to the powerful representation ability of graphs as well as the advances in graph mining and learning techniques. Our focus is anomaly detection on static graphs~\cite{akoglu2015graph}, where the graph structure and node features would are fixed overtime. Representative GAD can be roughly classified into four classes. Specifically, \textit{feature-based} methods~\cite{oddball,gao2006converting} extract hand-crafted features from the graph and use typical machine learning approaches to analyze those features. \textit{Proximity-based} methods~\cite{chen2013ascos} exploit the graph structure to measure the distances among nodes and anomalous nodes are assumed to have larger distances to other nodes. \textit{Community-based} methods~\cite{kuang2012symmetric} employ community detection approaches to cluster normal nodes. \textit{Relational-learning-based} methods~\cite{kang2011mining,koutra2011unifying} design graphical models to capture the relations among nodes and cast anomaly detection as a classification problem. Importantly, all these methods rely on the graph structure as an essential input. More recently, there are methods based on graph representation learning~\cite{10.1145/3340531.3411979,ReFeX}, whose key component is to learn the node embeddings via various technical as such graph neural networks.  We refer to \cite{akoglu2015graph} for a thorough discussion of these GAD methods.


\textbf{Adversarial graph analysis} Our work belongs to a long line of research that studies the adversarial robustness of various graph analysis tasks, such as node classification~\cite{Z_gner_2018,DBLP:journals/corr/abs-1902-08412,zhou2020robust}, link prediction~\cite{zhou2019attacking,waniek2019hide}, community detection~\cite{waniek2018hiding}, etc. For instance, \cite{zhou2019attacking} and \cite{waniek2019hide} investigated structural attacks against a wide range of similarity-based link prediction algorithms. \cite{Z_gner_2018} and \cite{DBLP:journals/corr/abs-1902-08412} initiated the study of the robustness of graph neural networks under both targeted and untargeted attacks. \cite{waniek2018hiding} investigated how an attacker can evade community detection tools in social networks. On the defence side, a lot of efforts have been devoted to enhancing the robustness of those analytic tools under attacks.
Representative approaches include adding regularization terms~\cite{jin2020graph} into the learning objective function, robust optimization~\cite{zhou2020robust}, injecting random noises to create ensembled classifiers~\cite{Z_gner_2019}, purifying the poisoned graphs~\cite{wu2019adversarial,10.1145/3336191.3371789}, and so on. All of the above works are task-specific and the developed methods cannot be easily applied to other tasks.

\section{Target GAD System}
\label{sec-GAD}
In this section, we introduce $\mathsf{OddBall}$~\cite{oddball} as the representative target GAD system of our attack. $\mathsf{OddBall}$ is an unsupervised feature engineering approach. 
Given a graph, $\mathsf{OddBall}$ extracts some carefully crafted features for each node in the graph, and computes an anomaly score for each node based on those features, where a larger score indicates that the node is more likely to be anomalous. 

\begin{figure}[h]
	\centering
	\begin{subfigure}[b]{0.23\textwidth}
		\centering
		\includegraphics[width=\textwidth,height=2.5cm]{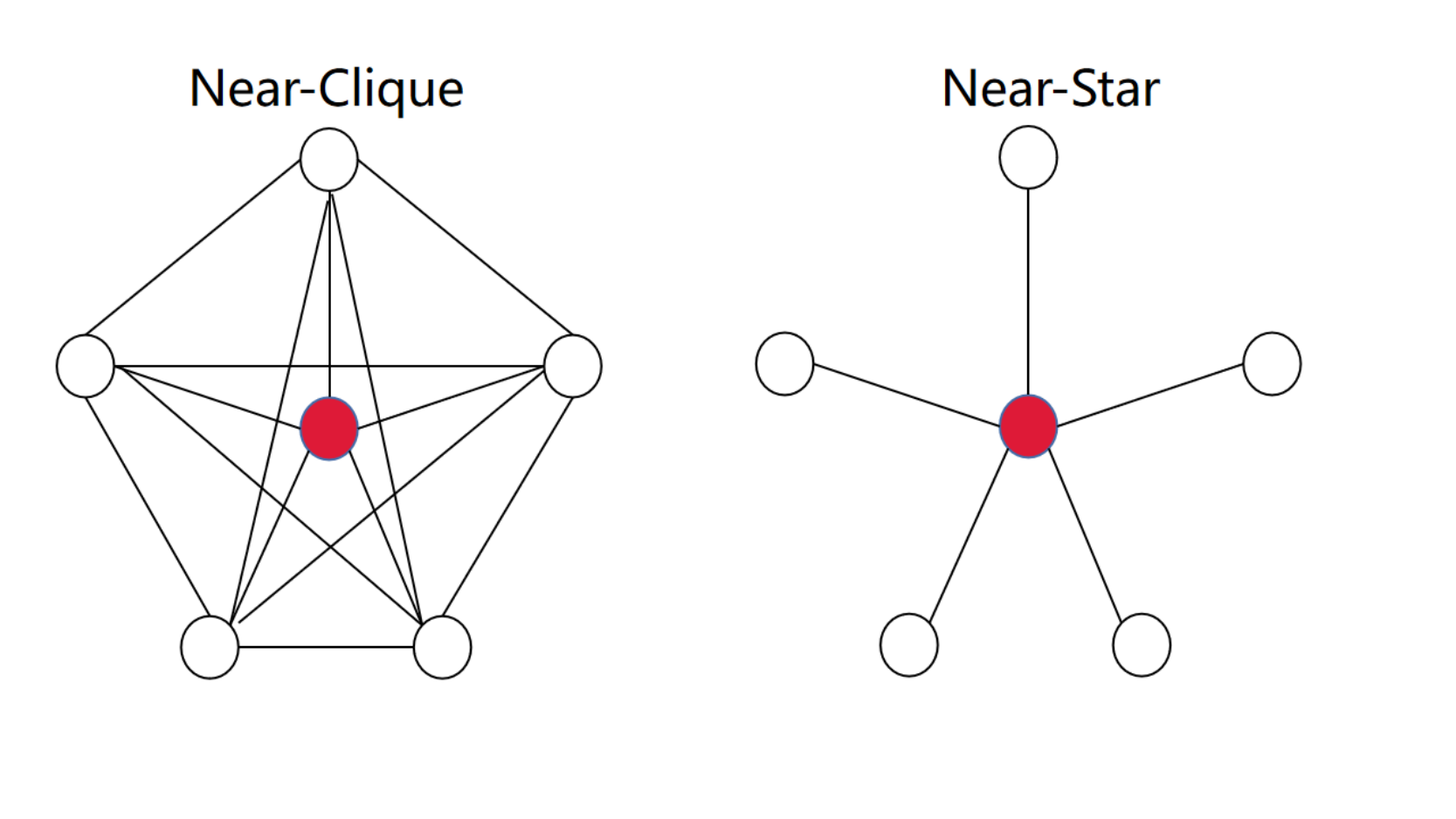}
		\caption{Anomalous patterns}
		\label{fig-anomalous-pattern}
	\end{subfigure}
	\hfill
	\begin{subfigure}[b]{0.23\textwidth}
		\centering
		\includegraphics[width=\textwidth,height=2.5cm]{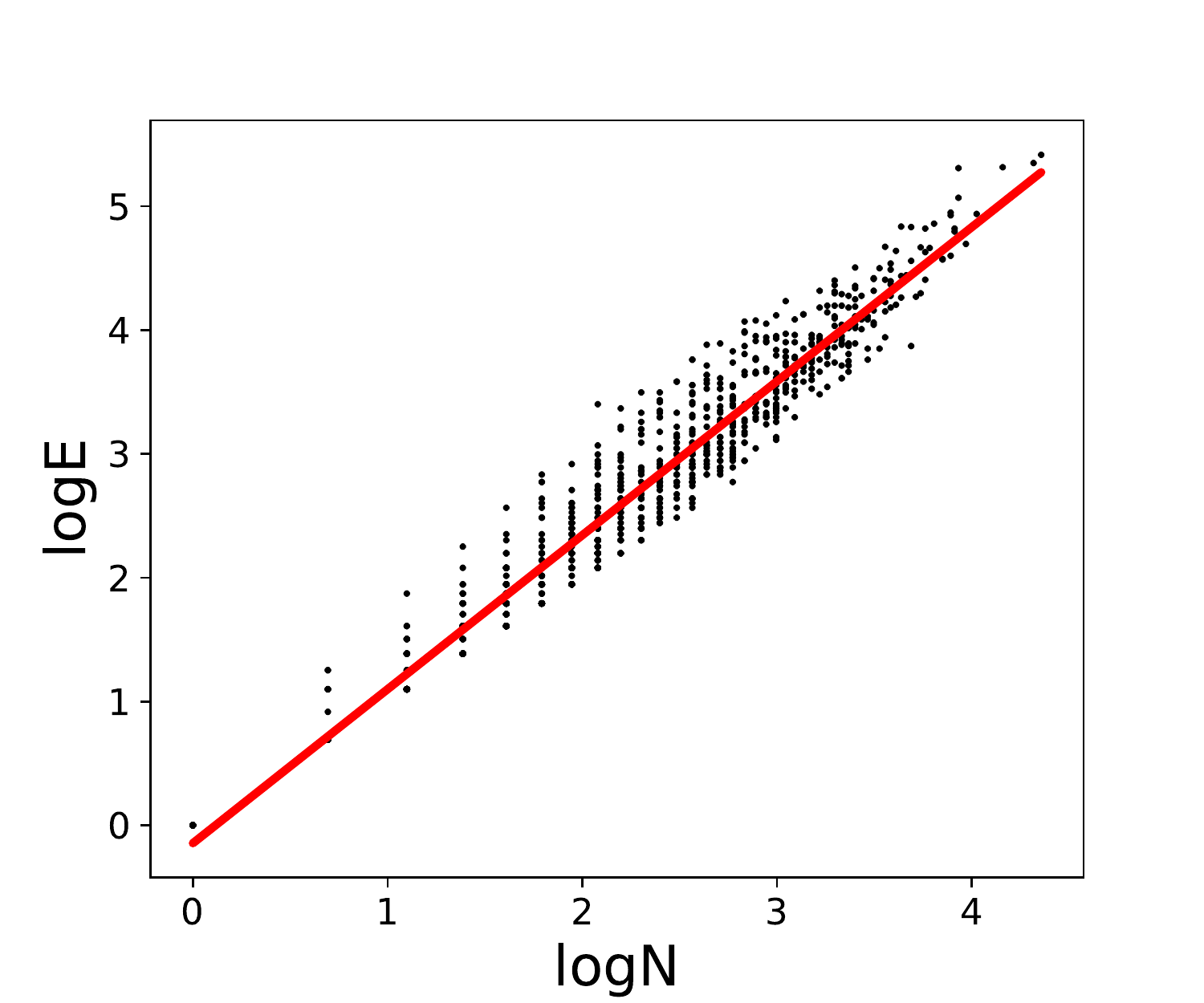}
		\caption{Feature distribution}
		\label{fig-feature}
	\end{subfigure}
\caption{(a) The \textit{near-clique} (left) and \textit{near-star} (right) structural patterns. (b) The distribution of node features and the regression line. The anomaly score is measured as the distance from a dot to the regression line along the vertical axis. }
\end{figure}

To formalize, we denote a graph as $\mathcal{G}=(V, E)$, where $V$ denotes a set of $n$ nodes and $E$ represents the edges. We consider a simple unweighted graph with the adjacency matrix $\mathbf{A} \in \{0,1\}^{n \times n}$.
$\mathsf{OddBall}$ focuses on the local structural information of the nodes for anomaly detection. Especially, for a node $v_i \in V$, $\mathsf{OddBall}$ examines an Egonet $\mathsf{ego}_i$ centered at $v_i$, where $\mathsf{ego}_i$ is the reduced sub-graph contains $v_i$ and its one-hop neighbors. 
An important finding of $\mathsf{OddBall}$ is that the Egonets for anomalous nodes tend to appear in either a \textit{near-clique} or \textit{near-star} structure (as shown in Fig.~\ref{fig-anomalous-pattern}). To detect such anomalous structures (i.e., \textit{near-clique} or \textit{near-star}), $\mathsf{OddBall}$ identifies two critical features $E_i$ and $N_i$ among others from the Egonet $\mathsf{ego}_i$, where $E_i$ and $N_i$ denote the number of edges and nodes in $\mathsf{ego}_i$, respectively. It was observed that $E_i$ and $N_i$ follow an \textit{Egonet Density Power Law}~\cite{oddball}: $E_i \propto N_i^\alpha, 1 \leq \alpha \leq 2$. The nodes that significantly deviate from this law are thus flagged as anomalous.


$\mathsf{OddBall}$ uses a regression-based approach to quantify the deviation. Specifically, let vectors $\mathbf{E} = (E_1,E_2, \cdots, E_n)$ and $\mathbf{N} = (N_1, N_2, \cdots, N_n)$ be the collected features of all nodes.  Based on the power law observation, one can use the following linear model to fit the pair of features $(E_i, N_i)$ for a node $i$:
\begin{equation}
\label{power law}
\ln E_{i}=\beta_{0}+\beta_{1} \ln N_{i}+\epsilon, 
\end{equation}
where the model parameters $\beta_{0}$ and $\beta_{1}$ are given by the Ordinary Least Square (OLS) \cite{Zdaniuk2014} estimation as:
\begin{equation}
\label{eqn-parameter}
[\beta_{0},\beta_{1}] = ([\mathbf{1},\ln \mathbf{N}]^{T}[\mathbf{1},\ln \mathbf{N}])^{-1}[\textbf{1},\ln \mathbf{N}]^{T}\ln \mathbf{E},
\end{equation}
where $\mathbf{1}$ is an $n$-dimensional vector of all $1$'s.

The anomaly score $S_i(\mathbf{A})$ for the node $v_i$ is then computed as
\begin{equation}
S_i(\mathbf{A})=\frac{\max(E_{i},e^{\beta_{0}}N_{i}^{\beta_{1}})}{\min(E_{i},e^{\beta_{0}}N_{i}^{\beta_{1}})}\ln(|E_{i}-e^{\beta_{0}}N_{i}^{\beta_{1}}|+1).
\end{equation}
Note that we made the dependency of $S_i(\mathbf{A})$ on $\mathbf{A}$ explicit as all of $\beta_0, \beta_1, E_i, N_i$  rely on $\mathbf{A}$.
As illustrated in Fig.~\ref{fig-feature}, the anomaly score $S_i(\mathbf{A})$ intuitively  measures the distance between the point $(N_{i},E_{i})$ and the regression line along the vertical axis. Finally, nodes with high anomaly scores (e.g., exceeding a certain threshold) are determined as anomalous by $\mathsf{OddBall}$.

\section{Formulation of Structural Poisoning Attacks}
\label{sec-formulation}
In this section, we formally formulate the structural poisoning attacks against $\mathsf{OddBall}$ \cite{oddball}.


\subsection{Threat Model}

\begin{figure}[h]
	\centering
	\includegraphics[width=0.45\textwidth,height=4cm]{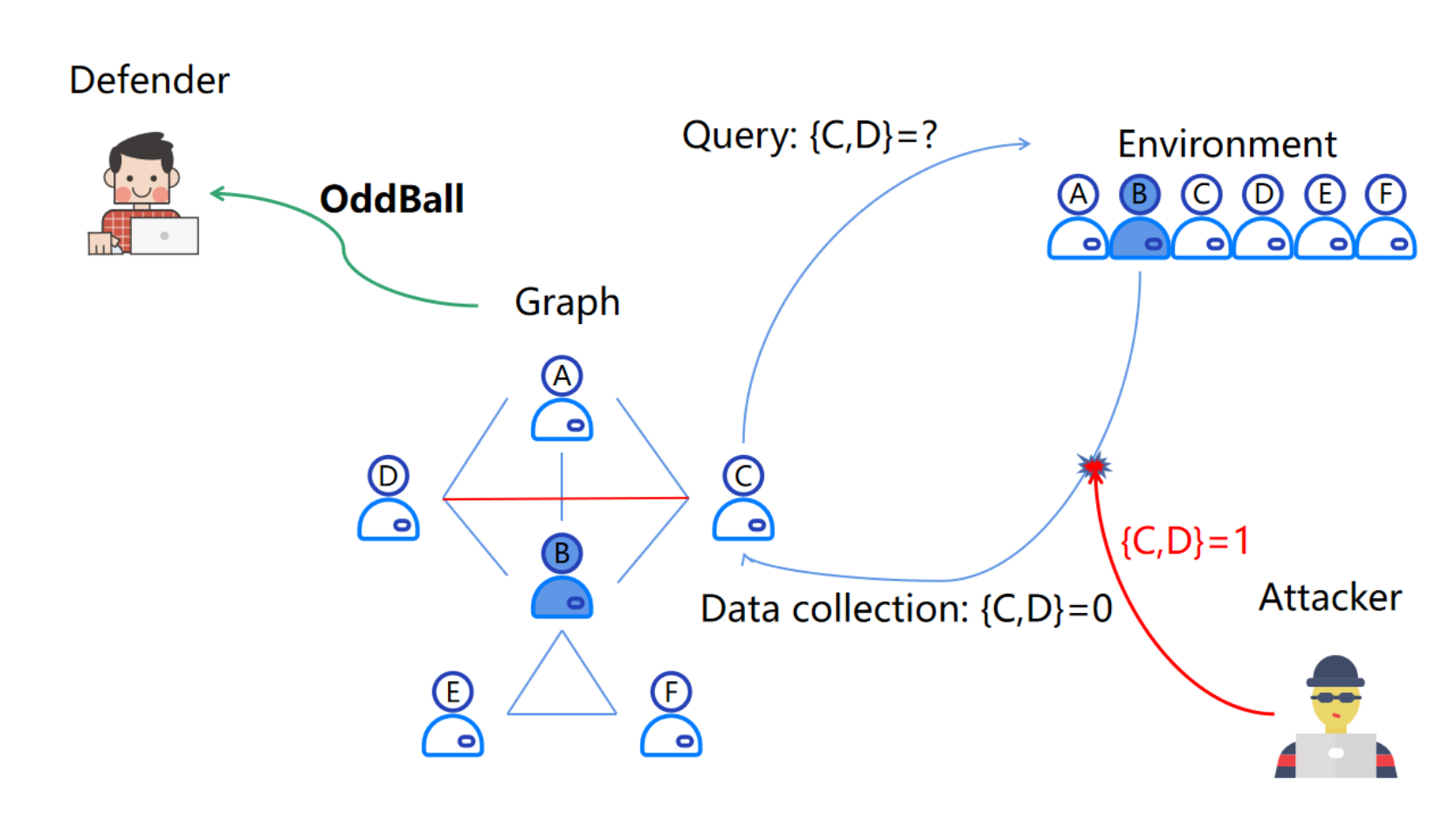}
	\caption{Interplay among defender, attacker and environment. Defender sends queries to the environment and constructs an observed network based on the query results. Meanwhile, an attacker can modify the query results, equivalently manipulating the network topology.} 
	\label{fig-interaction}
\end{figure}

We consider a system consisting of three parties: a defender, an attacker, and the outside environment,, the interplay among which is illustrated in Fig.~\ref{fig-interaction}. Specifically, a defender deploys $\mathsf{OddBall}$ to detect anomalous nodes in an existing network. In practice, the network is not readily available; instead, the defender will need to construct the network via data collection. We model data collections as a querying process where the defender sends queries consisting of pairs of nodes $(u,v)$ to the environment, which responds with the existence of the relation between $u$ and $v$ (i.e., whether there is an edge between $u$ and $v$). Then, based on the query results the defender can construct an observed network. This querying process widely models real-world scenarios such as taking surveys on friendships, conducting field experiments to measure the communication channels, etc.

An attacker, sitting between the defender and the outside environment, can tamper with the above data collection procedure by modifying the query results sent from the environment. For example, as shown in Fig.~\ref{fig-interaction}, the defender makes a query ``Is there a relation between $C$ and $D$?" to the environment; an attacker can change the query result ``$\{C,D\} = 0$" (non-existence) to ``$\{C,D\} = 1$" (existence), which is obtained by the defender. Consequently, the attacker is inserting an edge in the network constructed by the defender. That is, by modifying the query results, the attacker is equivalently manipulating the network topology, resulting in  structural attacks. We state the attacker's goal, knowledge, and capability as follows.

\begin{itemize}
	\item \textbf{Attacker's Goal.} We assume that the attacker has \textit{a set of target nodes} within the network, which are \textit{risky} nodes that have relatively high anomaly scores initially. The attacker's goal is thus to enable these target nodes to evade the detection. 
	
	
	\item \textbf{Attacker's Knowledge.} By Kerckhoffs's principle, we assume a worst-case scenario where the attacker has full knowledge of the network structure as well as the GAD system (i.e., $\mathsf{OddBall}$) deployed by the defender. That is, the attacker knows all the queries sent out by the defender as well as the results to all those queries.
	Later when considering transfer attacks, we relax this second assumption and attack other GAD systems in a black-box setting.

	\item \textbf{Attacker's Capability.} Again, we consider a worst-case scenario where the attacker can modify the query results at her choice. That is the attacker has control of the global structure of the graph and is able to add/delete edges from the graph. To further constrain the attacker's ability, we assume that the attacker can add/delete up to $B$ edges.
\end{itemize}

\subsection{Attack Problem}
We now formulate the attacks against GAD systems with $\mathsf{OddBall}$ as an instantiation.
We use $\mathcal{G}_0 = (V_0,E_0)$ to represent the ground-truth graph in the environment. An attacker, knowing the graph $\mathcal{G}_{0}$, has a set of target nodes $\mathcal{T} \subset V_0$,  and her goal is to reduce the probabilities that the nodes in $\mathcal{T}$ are detected as anomalous. To this end, the attacker is allowed to add or delete \textit{at most} $B$  edges in the graph $\mathcal{G}_{0}$ via tampering with the data collection process. As a result, a manipulated graph $\mathcal{G} = (V_0,E)$ is \textit{observed} by the defender, who will use a GAD system to detect anomalous nodes based on $\mathcal{G}$. We use $\mathbf{A}_0$ and $\mathbf{A}$ to denote the adjacency matrices of $\mathcal{G}_{0}$ and $\mathcal{G}$, respectively.

We emphasize that the graph manipulation process (i.e., changing $\mathcal{G}_0$ to $\mathcal{G}$) occurs before anomaly detection, resulting in a poisoning attack, which more suitably captures the nature of GAD in an unsupervised setting. Notably, the regression model (more precisely, $\beta_0$ and $\beta_1$) changes as the graph is modified. 
This also dramatically differentiates our structural attacks to previous evasion attacks where the detection model is fixed, and significantly complicates the design and realization of structural attacks.

To evade detection, the attacker tries to minimize the  anomaly scores of the target nodes in $\mathcal{T}$. In our formulation, we consider the weighted sum of scores, i.e., $S_{\mathcal{T}}(\mathbf{A}) = \sum_{i:v_i\in \mathcal{T}}\kappa_{i} S_i(\mathbf{A}) =  \sum_{i:v_i\in \mathcal{T}} \kappa_{i} \frac{\max(E_{i},e^{\beta_{0}^*}N_{i}^{\beta_{1}^*})}{\min(E_{i},e^{\beta_{0}^*}N_{i}^{\beta_{1}^*})}\ln(|E_{i}-e^{\beta_{0}^*}N_{i}^{\beta_{1}^*}|+1)$. For simplicity, in this paper, we consider the equal weight case, i.e., $\forall i, \kappa_{i}=1$, and our methods can be easily extended to the case with unequal weights. Then the attack can be formulated as the following optimization problem:


\begin{subequations}
	\label{eqn-opt-original}
	\begin{align}
		\mathbf{A}^* = &\argmin_{\mathbf{A}} S_{\mathcal{T}}(\mathbf{A}) \label{eq:goal}\\
		\text{s.t.} \quad &\beta_0^*, \beta_1^* = \mathsf{Regression} (\mathbf{A}), \label{eq:const1}\\
		&\frac{1}{2}||\mathbf{A}_0 - \mathbf{A}||_1 \leq B, \quad \mathbf{A} \in \{0,1\}^{n \times n} \label{eq:const2}
	\end{align}
\end{subequations}

where \eqref{eq:const1} denotes the function of deriving the parameters of the regression model from the graph $\mathbf{A}$, the constraint \eqref{eq:const2} ensures that the attacker can modify at most $B$ edges, and $\mathbf{A}^*$ is the optimal adversarial graph structure that the attacker aims to find.



To solve the optimization problem \eqref{eqn-opt-original}, we re-formulate it from several aspects. 
First, we note that the anomaly score $S_i(\mathbf{A})$ for node $v_i$ is a normalized distance to the regression line. To reduce the non-linearity and ease optimization, we omit the normalization term and use a proxy $\tilde{S}_i(\mathbf{A}) = \ln(|E_{i}-e^{\beta_{0}^*}N_{i}^{\beta_{1}^*}|+1)$ to approximate $S_i(\mathbf{A})$. Consequently, we will use an objective function $\tilde{S}_{\mathcal{T}}(\mathbf{A}) = \sum_{i:v_i\in \mathcal{T}} (E_{i}-e^{\beta_{0}^*}N_{i}^{\beta_{1}^*})^{2}$ acting as the surrogate of $S_{\mathcal{T}}(\mathbf{A})$ in the optimization process. We emphasize that $\tilde{S}_{\mathcal{T}}(\mathbf{A})$ is only used in the optimization process and we use the true anomaly scores $S_{\mathcal{T}}(\mathbf{A})$ for evaluation. Second, for OLS point estimation, the regression parameters $\beta_0^*$ and $\beta_1^*$ have closed-form solutions (Eqn.~\eqref{eqn-parameter}), allowing use to directly substitute $\beta_0^*$ and $\beta_1^*$ with functions of $\mathbf{A}$. 
At last, we can explicitly write the features $N_i$ and $E_i$ as 
$N_{i}=\sum_{j=1}^{n}\mathbf{A}_{ij}$, $E_{i}=N_{i}+\frac{1}{2}\mathbf{A}_{ii}^{3}.$ Finally, we can re-formulate the attack problem as:

\begin{subequations}
	\label{eqn-model}
	\begin{align}
		&\mathbf{A}^* = \argmin_{\mathbf{A}}\  \tilde{S}_{\mathcal{T}}(\mathbf{A}) \label{eq:goal2}\\
		&= \argmin_{\mathbf{A}} \sum_{i:i \in \mathcal{T}}(E_{i}-e^{(1,\ln N_{i})^{T}([\mathbf{1},\ln \mathbf{N}]^{T}[\mathbf{1},\ln \mathbf{N}])^{-1}[\textbf{1},\ln \mathbf{N}]^{T}\ln \mathbf{E}})^{2} \nonumber \\
		&\text{s.t.}\quad  N_{i}=\sum_{j=1}^{n}\mathbf{A}_{ij}, \quad E_{i}=N_{i}+\frac{1}{2}\mathbf{A}_{ii}^{3}; \label{eq:const21} \\
		&\quad \quad \frac{1}{2}||\mathbf{A}_0 - \mathbf{A}||_1 \leq B, \quad \mathbf{A} \in \{0,1\}^{n \times n} \label{eq:const22}.	 
	\end{align}
\end{subequations}

Solving the above discrete optimization problem leads to optimal a structural attack.
\section{Attack Methods}
\label{sec-methods}
In this section, we introduce three methods, $\mathsf{GradMaxSearch}$, $\mathsf{ContinuousA}$, and $\mathsf{BinarizedAttack}$, to solve the optimization problem \eqref{eqn-model}. Specifically, the first two proposed methods $\mathsf{GradMaxSearch}$ and $\mathsf{ContinuousA}$ are adapted from typical approaches in the literature for solving similar problems. We further propose $\mathsf{BinarizedAttack}$ to address their limitations.

\subsection{Conventional Methods}
Solving the optimization problem \eqref{eqn-model} to obtain the optimal graph structure is hard in general, mainly due to the integer variables involved. Thus a common approach is to relax the integral constraints, transforming the optimization problem to its continuous counterpart, for which gradient-descent-based optimization techniques could be employed. Ideally, a \textit{continuous} optimal solution $\tilde{\mathbf{A}}^*$ is obtained, which is then transformed to a \textit{discrete} solution  $\mathbf{A}^*$. 
This approach faces two central challenges: i) how to use the gradient information for the guidance of searching for $\tilde{\mathbf{A}}^*$, and ii) how to discretize $\tilde{\mathbf{A}}^*$ to obtain $\mathbf{A}^*$. Based on the previous works in the literature, we propose two methods. 


\subsubsection{$\mathsf{GradMaxSearch}$} 
Most of the previous works on structural attacks utilize a greedy strategy to solve the optimization problem in an iterative way. Specifically, the integer constraints on $\mathbf{A}$ are relaxed and in each iteration, the gradients of the objective with respect to each entry of $\tilde{\mathbf{A}}$ are calculated. Then, the entry with the largest gradient is picked for alternation (either deleting or adding an edge).
Intuitively, a larger gradient indicates a bigger impact on the objective value. The iteration continues until budget constraint $B$ is reached. We implement this idea and term the resulting algorithm as $\mathsf{GradMaxSearch}$.

Regarding the implementation of $\mathsf{GradMaxSearch}$, we note that when picking the entry associated with the largest gradient, one should pay attention to the \textit{signs of the gradients}. For example, when $\mathbf{A}_{ij}=0$ we need to ensure that the corresponding gradient $\frac{\partial AS(v_{a})}{\partial \mathbf{A}_{ij}}<0$ (and vice versa) to make the operation (add or delete) valid. In addition, to avoid repeatedly adding and deleting the same edge, we maintain a pool to record the edges that have not been modified. Meanwhile, we also avoid the operations that would result in singleton nodes. 

\subsubsection{$\mathsf{ContinuousA}$}
An apparent shortcoming of $\mathsf{GradMaxSearch}$ is that the objective function is only optimized through $B$ steps, where $B$ is the attacker's budget. An alternative way is thus to totally treat $\mathbf{A}$ as continuous variables $\tilde{\mathbf{A}} \in [0,1]^{n\times n}$ and optimize the objective function until it converges. We term this method as $\mathsf{ContinuousA}$.


In detail, by solving the optimization problem in the continuous domain using gradient-descent, we obtain the sub-optimal continuous solution $\tilde{\mathbf{A}}^*$. We then calculate the differences in absolute values between the original $\mathbf{A}$ and $\tilde{\mathbf{A}}^*$, and pick those edges associated with the top-$B$ absolute differences to modify. 

\subsection{$\mathsf{BinarizedAttack}$}

\begin{algorithm}[t]
	\caption{$\mathsf{BinarizedAttack}$}
	\label{alg-binarizedattack}
	\textbf{Input}: clean graph $\mathbf{A}^{0}$, anomaly score function $AS$,  target set $V=\{v_{a}\}_{a=1}^{\tau}$, budget $B$, hyper-parameter set $\Lambda=\{\lambda_{k}\}_{k=1}^{K}$, iteration number $T$, learning rate $\eta$.\\
	\textbf{Parameter}: Perturbation $\dot{\mathbf{Z}}$.\\
	\begin{algorithmic}[1] 
		\STATE Let $t=0$ and initialize $\dot{\mathbf{Z}}$.
		\FOR {$k\leftarrow 1,2,...,K$}
		\WHILE{$t\leq T$}
		\STATE \textbf{Forward Pass}:
		\STATE \quad Calculate $\mathbf{Z} =- \mathsf{binarized}(2\cdot \dot{\mathbf{Z}}-1).$
		\STATE \quad Calculate $\mathbf{A}=(\mathbf{A}_0 - 0.5 \cdot \mathbf{1}^{n\times n})\odot \mathbf{Z}+0.5$.
		\STATE \quad Calculate $N_{i}=\sum_{j=1}^{n}\mathbf{A}_{ij}$, $E_{i}=N_{i}+\frac{1}{2}\mathbf{A}_{ii}^{3}.$
		\STATE \quad Obtain goal function $\mathcal{L}(\{v_{a}\}_{a=1}^{\tau})$ according to \ref{eq:goal3}.
		\STATE \textbf{Backward Pass}:
		\STATE \ \ \ \ $\forall i,j\in{1,2,...,n},$ calculate the gradient of the goal function $\mathcal{L}(\{v_{a}\}_{a=1}^{\tau})$ w.r.t $\dot{\mathbf{Z}}_{ij}$, i.e., $\frac{\partial \mathcal{L}(\{v_{a}\}_{a=1}^{\tau})}{\partial \dot{\mathbf{Z}}_{ij}}.$
		\STATE \textbf{Projection Gradient Descent}:
		\STATE \ \ \ \ $\dot{\mathbf{Z}}\rightarrow\prod_{[0,1]}(\dot{\mathbf{Z}}-\eta\frac{\partial \mathcal{L}(\{v_{a}\}_{a=1}^{\tau})}{\partial \dot{\mathbf{Z}}_{ij}})$
		\ENDWHILE
		\STATE \textbf{return} $\dot{\mathbf{Z}}$
		\ENDFOR
		\FOR {$b\leftarrow 1,2,...,B$}
		\STATE Pick out $\dot{\mathbf{Z}}=min \ \mathcal{L}(\{v_{a}\}_{a=1}^{\tau})$ satisfies $\sum \mathbf{Z}=-b$.
		\STATE Get poisoned graph $\mathbf{A}^{b}=(\mathbf{A}_0 - 0.5 \cdot \mathbf{1}^{n\times n})\odot \mathbf{Z}+0.5$.
		\STATE \textbf{return} $\mathbf{A}^{b}$.
		\ENDFOR
	\end{algorithmic}
\end{algorithm}
Both $\mathsf{GradMaxSearch}$ and $\mathsf{ContinuousA}$ have their own limitations. For
$\mathsf{GradMaxSearch}$, one major limitation is that the gradient only indicates a relatively
\textit{small fractional} update on the corresponding entry in $\mathbf{A}$; while a discrete value ($\pm 1$) is
actually updated. This would not necessarily optimize the objective. Moreover, due
to budget constraint $B$, the objective is only optimized through $B$ steps. $\mathsf{ContinuousA}$ treating $\mathbf{A}$ as continuous in the whole process of optimization, totally ignoring the effect of discrete updates on the objective function. Furthermore, without careful design, converting from the fractional optimal solution $\tilde{\mathbf{A}}^*$ to $\mathbf{A}^*$ may lead to arbitrary bad performances.



We propose $\mathsf{BinarizedAttack}$ to mitigate these limitations. At a high level, $\mathsf{BinarizedAttack}$ is a gradient-descent-based approach that optimizes the objective in iterations. Each iteration consists of a forward pass, where the objective function is evaluated on some decision variables, and a backward pass, where the decision variables are updated based on calculated gradients.  The core idea of $\mathsf{BinarizedAttack}$ lies in the design of two sets of decision variables as well as the way of utilizing gradient information. Specifically, we associate each entry $\mathbf{A}_{ij}$ with a \textbf{discrete dummy} (the use of \textit{dummy} will be explained later) decision variable $\mathbf{Z}_{ij}\in \{-1,+1\}$, where $\mathbf{Z}_{ij}=-1$ indicates that the corresponding entry $\mathbf{A}_{ij}$ will be modified and vice versa. For example, if $\mathbf{A}_{ij} = 1$ and $\mathbf{Z}_{ij}=-1$, we will change $\mathbf{A}_{ij}$ to $0$. Let $\mathbf{A}_0$ and $\mathbf{A}$ be the original and modified adjacency matrix, respectively, which are connected through the decision variables $\mathbf{Z}$ by
\begin{equation}
	\mathbf{A}=(\mathbf{A}_0 - 0.5 \cdot \mathbf{1}^{n\times n})\odot \mathbf{Z}+0.5,
\end{equation}
where $\odot$ denotes element-wise multiplication between two matrices. 

We further associate each entry $\mathbf{A}_{ij}$ with a \textbf{continuous soft} decision variable $\dot{\mathbf{Z}} \in [0,1]$ to facilitate gradient computation. The two sets of decision variables $\dot{\mathbf{Z}}$ and $\mathbf{Z}$ are related by
\begin{align}
\label{eqn-z-z}
 \mathbf{Z} =- \mathsf{binarized}(2\cdot \dot{\mathbf{Z}}-1),
\end{align}
where we define the function $\mathsf{binarized}(x)$ as $\mathsf{binarized}(x) = + 1$ if $x\geq 0$ and $\mathsf{binarized}(x) = - 1$ otherwise.
Since our objective function $\tilde{S}_{\mathcal{T}}(\mathbf{A})$ depends on $\mathbf{A}$, we can easily rewrite it as a function relying on the decision variables $\dot{\mathbf{Z}}$ and $\mathbf{Z}$ as $\tilde{S}_{\mathcal{T}}(\dot{\mathbf{Z}},\mathbf{Z})$.

We proceed to handle budget constraint \eqref{eq:const22}. Our goal is to transform this constraint as part of the objective function so that we can thoroughly optimize the objective beyond $B$ steps. To this end, we impose a LASSO penalty~\cite{Tibshirani94regressionshrinkage} on the continuous soft decision variables $\dot{\mathbf{Z}}$. Our choice of LASSO comes from the fact that LASSO can obtain sparser solutions compared with the L2 penalty~\cite{Hoerl1}. Based on Eqn~\eqref{eqn-z-z}, we can observe that a larger $\dot{\mathbf{Z}}_{ij}$ indicates that it is more likely to modify the entry $\mathbf{A}_{ij}$ (i.e., $\mathbf{Z}_{ij} = -1$). As a result, in the optimization process, while the LASSO penalty term pushes the entries in $\dot{\mathbf{Z}}$ to zero, it is also restricting the modifications made to $\mathbf{A}$, achieving a similar effect of the budget constraint.

Now, we can reformulate the attack problem as an optimization problem with $\dot{\mathbf{Z}}$ and $\mathbf{Z}$ as decision variables:
\begin{subequations}
	\label{bin-model}
	\begin{align}
	&\mathbf{\dot{\mathbf{Z}}}^* = \argmin_{\mathbf{\dot{Z}}} \sum_{a=1}^{\tau}(E_{a}-e^{\rho})^{2}+\lambda||\dot{\mathbf{Z}}||_{1}^{1} \label{eq:goal3}\\
	&\text{s.t.}\quad \rho = (1,\ln N_{a})^{T}([1,\ln N]^{T}[1,\ln N])^{-1}[1,\ln N]^{T}\ln E \\
	&\quad \quad  N_{i}=\sum_{j=1}^{n}\mathbf{A}_{ij}, \quad E_{i}=N_{i}+\frac{1}{2}\mathbf{A}_{ii}^{3}; \label{eq:const31} \\
	&\quad \quad \mathbf{A}=(\mathbf{A}^{0}-0.5)\odot \mathbf{Z}+0.5; \label{eq:const32}. \\
	&\quad \quad \mathbf{Z}=-\mathsf{binarized}(2\cdot{\mathbf{Z}}-1). \label{eq:const33} 
	\end{align}
\end{subequations}
Note that we used a parameter $\lambda$ to tune the relative importance of the adversarial objective and the penalty term.

Now, we can solve \eqref{bin-model} through iteration. Specifically, in the forward pass, we will evaluate the objective by the discrete dummy variables $\mathbf{Z}$. Intuitively, this will more accurately measure the effect of discrete updates of the graph structure on the objective. In the backward pass, we can compute the gradients with respect to $\dot{\mathbf{Z}}$, and update $\mathbf{Z}$ from~\eqref{eq:const33}. 
In this way, by observing $\dot{\mathbf{Z}}$, we can obtain the entries in $\mathbf{A}$ that the attacker needs to modify. We note that the discrete variables $\mathbf{Z}$ are only used to evaluate the objective function for optimization; the final decisions are made from the soft variables $\dot{\mathbf{Z}}$, thus the reason why $\mathbf{Z}$ are called dummy variables. The pseudo-code of $\mathsf{BinarizedAttack}$ is presented in Alg.~\ref{alg-binarizedattack}.



\section{Attack Transferability}
\label{sec-transfer}
In this section, we investigate the attack transferability of $\mathsf{BinarizedAttack}$ to two other GAD systems that are based on representation learning.

\subsection{Representation-learning-based GAD}
Recently, graph analysis based on representation learning has attracted extensive research attention. Graph representation learning~\cite{hamilton2020graph} aims to learn a low-dimensional latent vector for each node (called embedding) that could capture the feature as well as the structural information of the graph. Those learned embeddings are then used in a wide range of down-stream tasks such as node classification and link prediction. Following this trend, graph representation learning has also been used as the core technique for anomaly detection. These techniques entitle the defender the freedom to chooce different GAD systems in practice, which raises a practical question: could an attack method that is specifically designed for a target GAD system be effective to other GAD systems? We investigate this problem of attack transferability by using $\mathsf{BinarizedAttack}$ to attack two other GAD systems (GAL~\cite{10.1145/3340531.3411979} and ReFeX~\cite{ReFeX}) based on graph representation learning.

At a high level, representation-learning-based GAD systems have two components: representation learning and classification. First, given a graph as input, the system learns the embeddings of nodes using various methods (e.g., graph neural networks, random walk, etc.). These embeddings are then fed into classifiers such as Multi-Layer Perceptron (MLP) which will classify nodes as anomalous or benign. These GAD systems differ mainly in the methods used for learning the node embeddings. We focus on two representative systems as below.

\subsubsection{GAL~\cite{10.1145/3340531.3411979}} GAL utilizes Graph Neural Networks \cite{kipf2017semisupervised} (GNNs) to learn the node embeddings. In particular, GAL replaced the loss function in typical GNNs with a \textit{graph anomaly loss} that is specially designed for anomaly detection. It is a class-distribution-aware margin loss which solves the imbalanced problem in anomaly detection via automatically adjusting the margins for the minority class (i.e., anomaly). Specifically, the loss on node $u$ is computed as:

\begin{align}
\mathcal{L}(u)=&E_{u_{+}\sim \mathcal{U}_{u+},u_{-}\sim \mathcal{U}_{u-}} max\{0,g(u,u_{-})-g(u,u_{+})+\Delta_{y_{u}}\} \nonumber \\
&\text{where} \quad \Delta_{y_{u}}=\frac{C}{n_{y_{u}}^{1/4}}. \label{eqn-gal}
\end{align}
Here $\mathcal{U}_{u+}$ denotes the set of nodes share the same label with $u$, and vice versa. $g(u,u^{\prime})=f(u)^{T}f(u^{\prime})$ measures the similarity of the representation of two nodes $u$ and $u^{\prime}$, $f$ is the GNN. $C$ is the constant hyperparameter to be tuned. $\Delta_{y_{u}}$ is proved to best weigh up between the improvement of the generalization of minority class and the sub-optimal margin for the frequent class.


\subsubsection{ReFeX~\cite{ReFeX}}
ReFeX (Recursive Feature eXtraction) is a novel algorithm that combines node-based local features with neighborhood (ego-based) features recursively, and output regional features which can capture the behavior information in large social networks. ReFeX aggregates the local and ego-based features and use them to create recursive features. Local features are essentially node degree measures. Egonet features are $\mathbf{N}$ and $\mathbf{E}$ as mentioned in Section III. ReFeX recurses the local and ego-based features by calculating their means and sums. Next ReFeX implements the pruning method using \textit{vertical logarithmic binning}. At last, ReFeX transforms the pruned recursive features to binary-valued embeddings. The pruned recursive features proved to efficiently capture the inner structural information provided by the graph. In this paper, we make use of these features for node anomaly detection.

\subsection{Transfer attack methodology}
\label{sec-transfer-method}
Our transfer attack to both GAL and ReFeX consists of four steps: \textit{data pre-processing}, \textit{targets identification}, \textit{graph poisoning} and \textit{evaluation}. 

\subsubsection{Data pre-processing}
$\mathsf{OddBall}$ operates in an unsupervised setting while both GAL and ReFeX are supervised methods. We thus pre-process the data by assigning anomaly labels to a set of nodes. Specifically, given a graph, we first use $\mathsf{OddBall}$ to compute the anomaly scores for all the nodes and label those nodes with high anomaly scores as anomalous. We then randomly split the nodes into training and testing sets.
\subsubsection{Targets identification}
As $\mathsf{BinarizedAttack}$ is a targeted attack, the goal of our transfer attack is not to decrease the prediction accuracies of the GAD systems. Instead, we are interested in changing the prediction results of a set of nodes -- those nodes that are initially predicted as anomalous by the GAD systems.
Thus, we feed the pre-process data into GAL or ReFeX and obtain the predicted labels of the test nodes and pick those nodes \textit{predicted as anomalous} as our attack targets. 
\subsubsection{Graph poisoning} With the targeted nodes identified from the previous step, we directly use $\mathsf{BinarizedAttack}$ to generate a poisoned graph. We emphasize that the attack occurred in a black-box setting, where we do not need any information from GAL nor ReFeX.

\subsubsection{Evaluation} We provide the clean graph and poisoned graph separately as input to GAL and ReFeX. We evaluate the performance of transfer attack mainly from two aspects: 1) whether the target nodes can successfully evade the detection of GAL or ReFeX; 2) the changes of node embeddings generated by GAL and ReFeX before and after attack. 

We present the detailed experiment settings and results in Section~VIII.

\section{Possible Countermeasures}
\label{sec-defense}
Defense approaches against data poisoning attacks have been extensive studied mainly in the computer vision domain. Typical methods include sanitizing the dataset by identifying and removing the maliciously injected data points, training robust models that could still perform well on poisoned data, and so on. More recently, there are also works that explore defense approaches in the graph learning domain. 
For example, based on the observation that $\mathsf{netattack}$ \cite{Z_gner_2018} tends to increase the rank of the adjacency matrix, \cite{10.1145/3336191.3371789} proposed the low-rank approximation of the adjacency matrix by SVD for defense. Graph variational autoencoder is used in \cite{DBLP:journals/corr/abs-2006-08900} to generate a smooth adjacency matrix to defend two popular attacks $\mathsf{metattack}$\cite{DBLP:journals/corr/abs-1902-08412} and $\mathsf{netattack}$. A robust GCN~\cite{kipf2017semisupervised} model is proposed in \cite{zhu2019robust} by formulating the Gaussian distribution based on the hidden features of GCN and penalizing the high variance to enhance the robustness. \cite{DBLP:journals/corr/abs-1908-07558} and \cite{DBLP:journals/corr/abs-2006-08149} introduced the attention mechanism to decrease the effect of adversarial links in the attacked graph.

In this paper, we further explore possible countermeasures to defend against $\mathsf{BinarizedAttack}$. Our key observation is that, since OLS \cite{Zdaniuk2014} estimation  is sensitive to poisoned data in the training process, we can use robust estimators  instead to estimate the regression parameters. Huber \cite{10.1214/aoms/1177703732} proposed the Huber loss function to replace the mean square loss:
\begin{equation}
\rho_{Huber}(t)=\left\{
\begin{aligned}
&\frac{1}{2}t^{2},\quad  \text{if} \ |t|\leq k \\
&k|t|-\frac{1}{2}k^{2}, \quad \text{if} \ |t| > k,
\end{aligned}
\right.
\end{equation}
where $k$ is the hyper-parameter to penalize the outliers linearly instead of quadratically.

Random Sample Consensus (RANSAC \cite{ransac}) uses Huber loss with $k=1$ for robust estimation.
It was known that RANSAC is able to estimate the model parameters that achieve high accuracy even when the dataset contains a significant number of outliers. In light of this, we adopt the robust estimation method from  RANSAC to estimate the parameters of $\mathsf{OddBall}$ (thus a robust version of $\mathsf{OddBall}$) from the poisoned graph generated by $\mathsf{BinarizedAttack}$. Our experiments show that this robust estimation approach can slightly mitigate the attack effect. The defense of structural poisoning attack remains as an important future work.

\section{Experiments}
\label{sec-exp}


\subsection{Datasets and Experiment Settings}
We evaluate the algorithms on two synthetic and three real-world datasets:

\subsubsection{Synthetic datasets} We use two random graph models, Erdos-Renyl (\textbf{ER})~\cite{erdHos1960evolution} and Barabasi-Albert (\textbf{BA})~\cite{Albert_2002}, to generate graphs with a node number of $1000$. For \textbf{ER} graphs, we set the link probability as $0.02$. We create \textbf{BA} graphs with $m = 5$, where $m$ is the number of edges attaching from a new node to existing nodes.

\subsubsection{Real-world datasets} \textbf{Blogcatalog}~\cite{zafarani2014users} is a social network indicating follower/followee relationships among bloggers in a blog sharing website. The network has 88,800 nodes and 2.1M edges. \textbf{Wikivote}~\cite{leskovec2010predicting} contains all the Wikipedia voting data from the inception of Wikipedia till January 2008. Nodes in the network represent Wikipedia users and the edge represents that user $i$ voted on user $j$. The dataset contains around 7000 nodes and 0.1M edges. \textbf{Bitcoin-Alpha}~\cite{kumar2018rev2} is a who-trusts-whom social network of traders who trade Bitcoin on the \textbf{Bitcoin-Alpha} platform. It is a weighted signed network with weights ranging from $+10$ to $-10$. It has more than 3000 nodes and 24186 edges. We pre-process the data by removing all the negative edges and erasing the weights of all the remaining edges, resulting
in an unsigned unweighted version of \textbf{Bitcoin-Alpha}. For all the three
real datasets, we randomly sample the connected sub-graph with around $1000$ nodes from the whole graph. The
statistics of the real datasets are summarized in Table~\ref{table-datasets}.

 \begin{table}[h]
	\centering
	\caption{Statistics of datasets}
	\label{table-datasets}
	\resizebox{0.9\columnwidth}{!}{%
	\begin{tabular}{|c|c|c|c|c|c|}
		\hline
		\hline
		& \textbf{ER} & \textbf{BA} & \textbf{Blogcatalog} & \textbf{Wikivote} & \textbf{Bitcoin-Alpha}\\
		\hline
        \# of nodes & 1000& 1000& 1000& 1012& 1025\\
        \hline
        \# of edges & 9948& 4975& 6190& 4860& 2311\\
        \hline
	\end{tabular}
}
\end{table}

\subsubsection{Settings}
For all datasets we determine our target set by sampling 10 or 30 target nodes from the top-$50$ nodes based on $\mathsf{AScore}$ (From now on we denote anomaly score as $\mathsf{AScore}$ for simplicity) rankings. For each experiment we sample target nodes 5 times individually and report the mean values to measure the efficacy of the structural attack on $\mathsf{OddBall}$. For structural attack (add/delete edges) on node anomaly detection, we aim to minimize the $\mathsf{AScore}$ sums for target set $V$ under different budget constraint. We determine the max budget $B$ for each case by the convergence of $\mathsf{AScore}$ for the three structural attack. We stop attacking until the changes of $\mathsf{AScore}$ saturated.


\subsection{Attack Performance of $\mathsf{BinarizedAttack}$}

\begin{figure}
	\centering
	\begin{subfigure}[b]{0.234\textwidth}
		\centering
		\includegraphics[width=\textwidth,height=3cm]{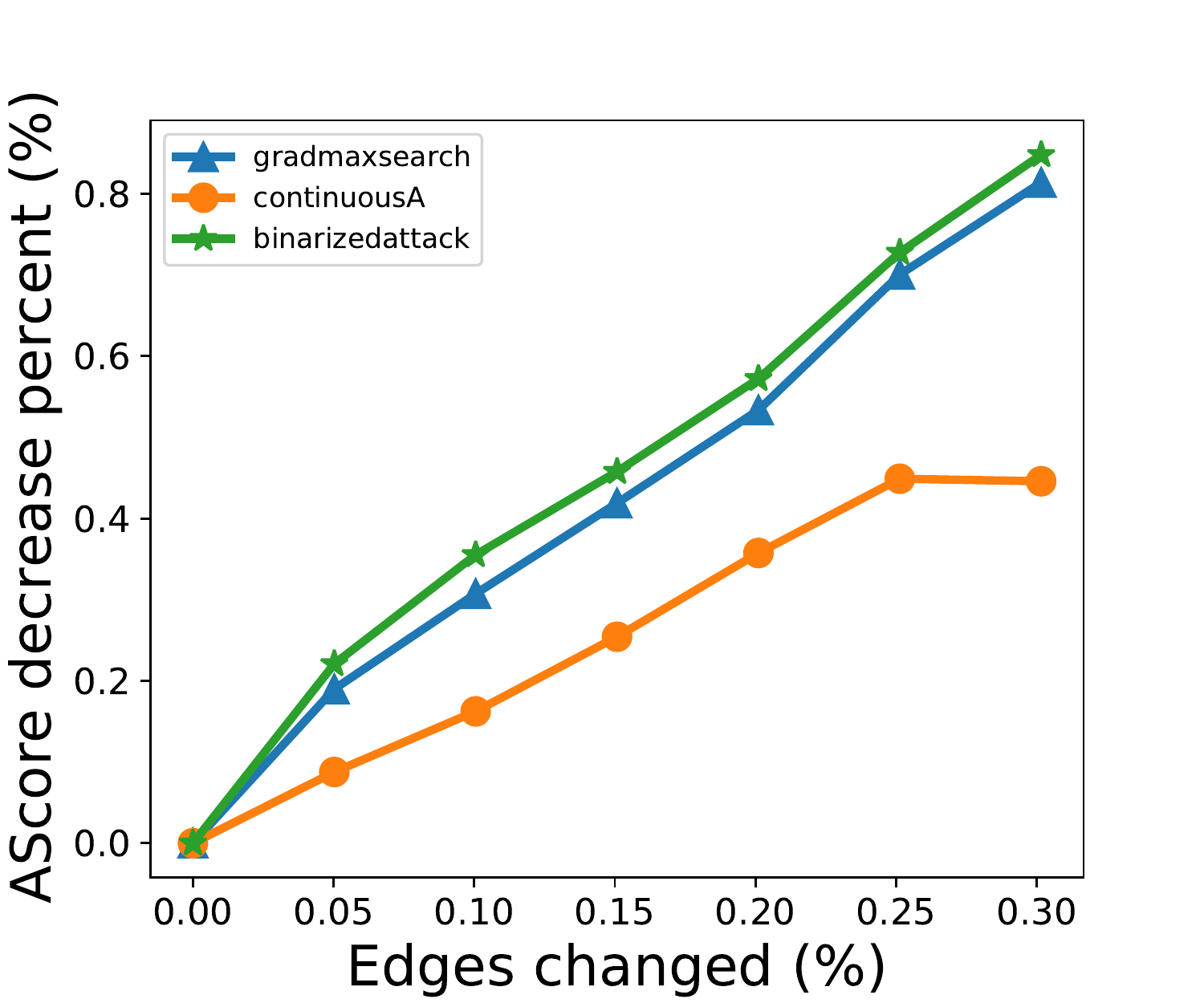}
		\caption{\textbf{ER} graph}
	\end{subfigure}
	\hfill
	\begin{subfigure}[b]{0.234\textwidth}
		\centering
		\includegraphics[width=\textwidth,height=3cm]{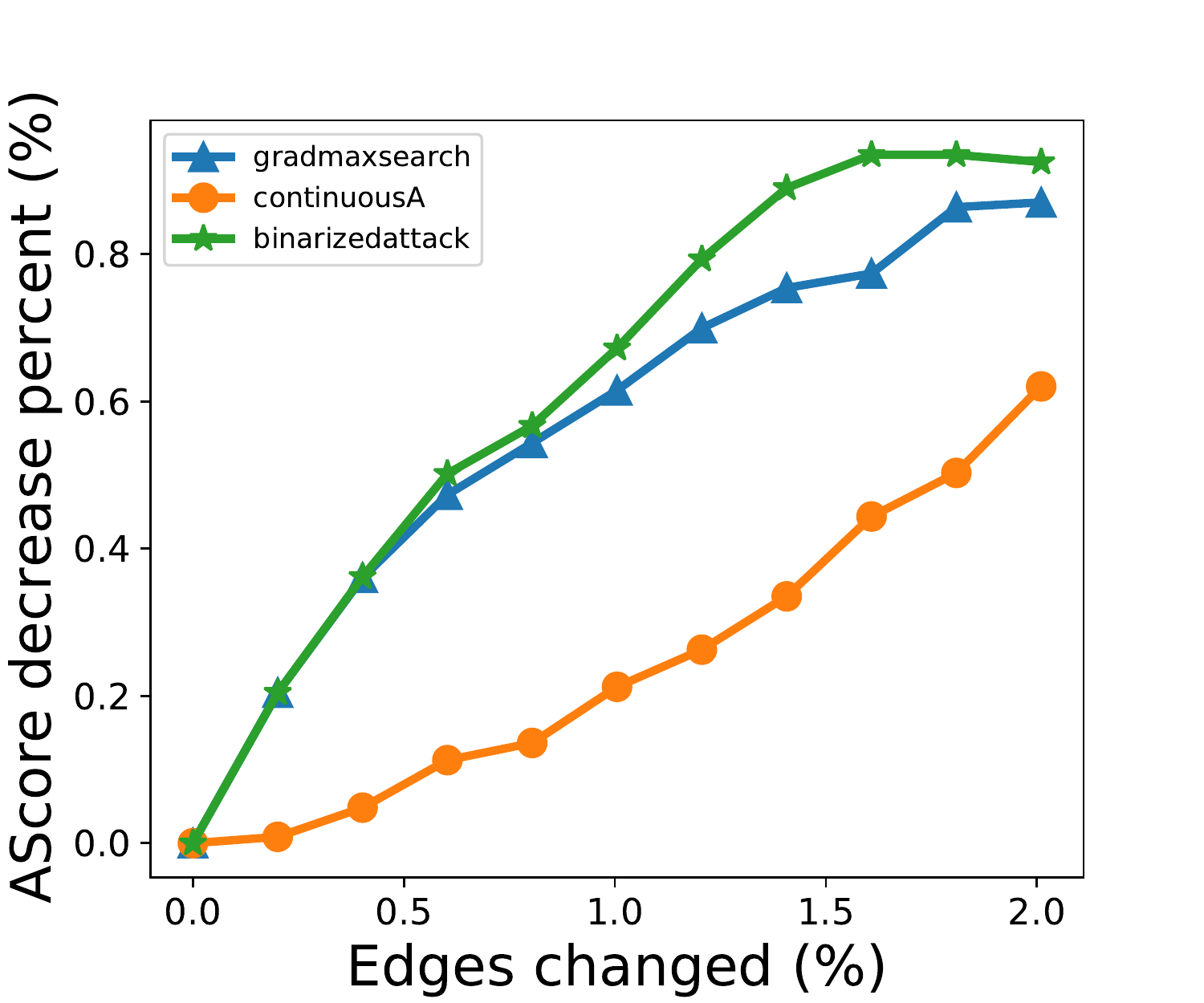}
		\caption{\textbf{BA} graph}
	\end{subfigure}
	\hfill
	\begin{subfigure}[b]{0.234\textwidth}
		\centering
		\includegraphics[width=\textwidth,height=3cm]{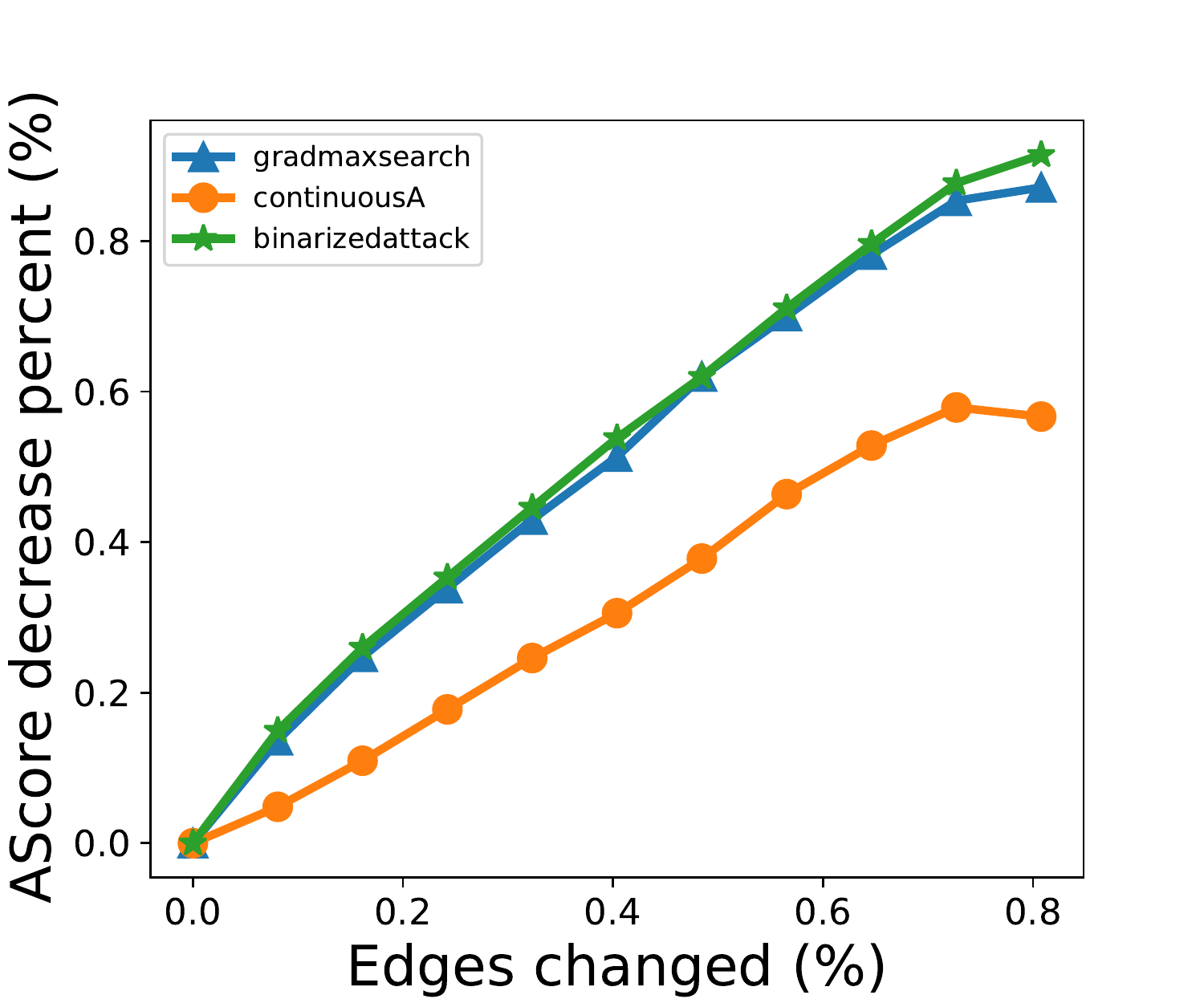}
		\caption{\textbf{Blogcatalog}-10}
	\end{subfigure}
	\hfill
	\begin{subfigure}[b]{0.234\textwidth}
		\centering
		\includegraphics[width=\textwidth,height=3cm]{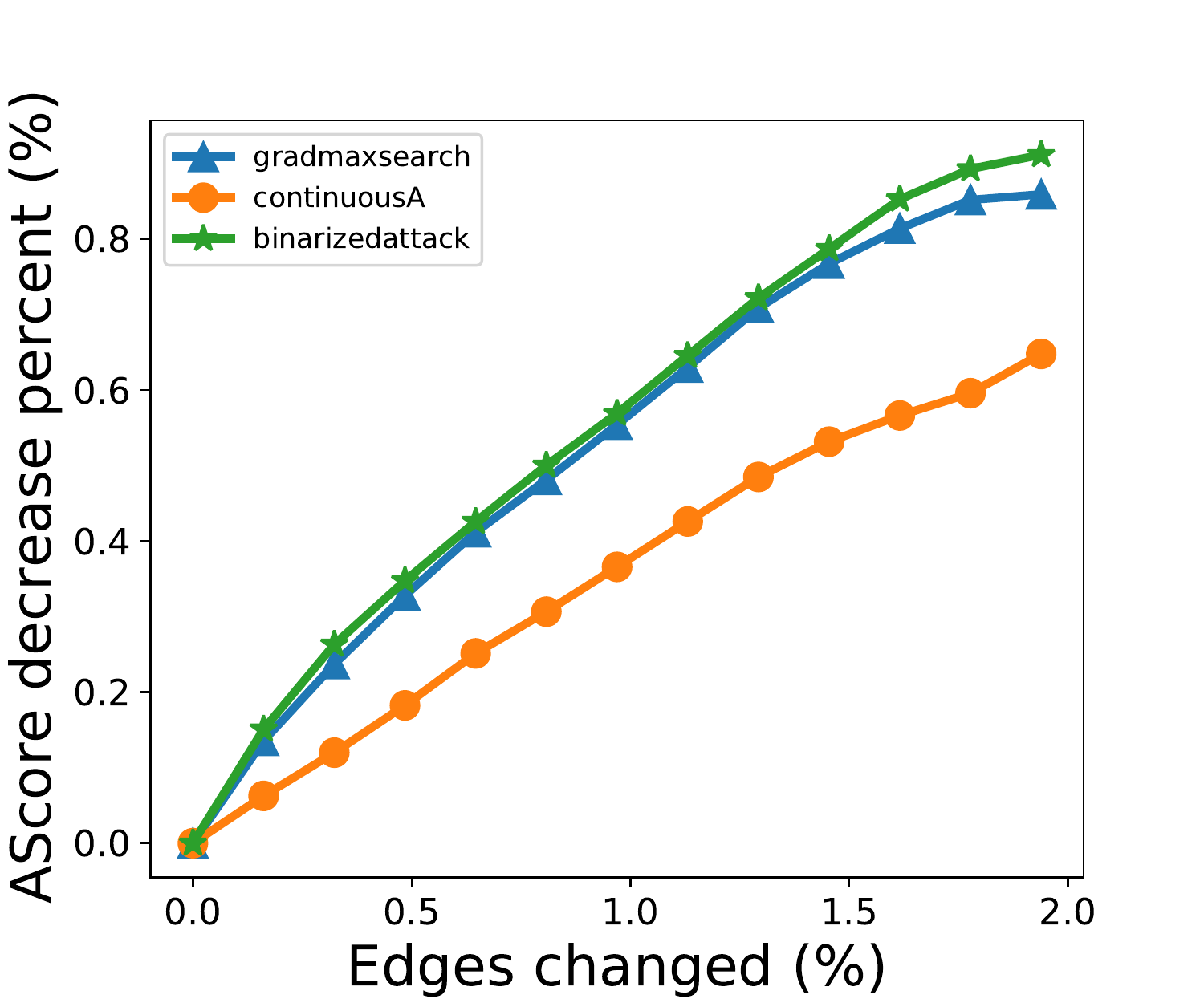}
		\caption{\textbf{Blogcatalog}-30}
	\end{subfigure}
	\hfill
	\begin{subfigure}[b]{0.234\textwidth}
		\centering
		\includegraphics[width=\textwidth,height=3cm]{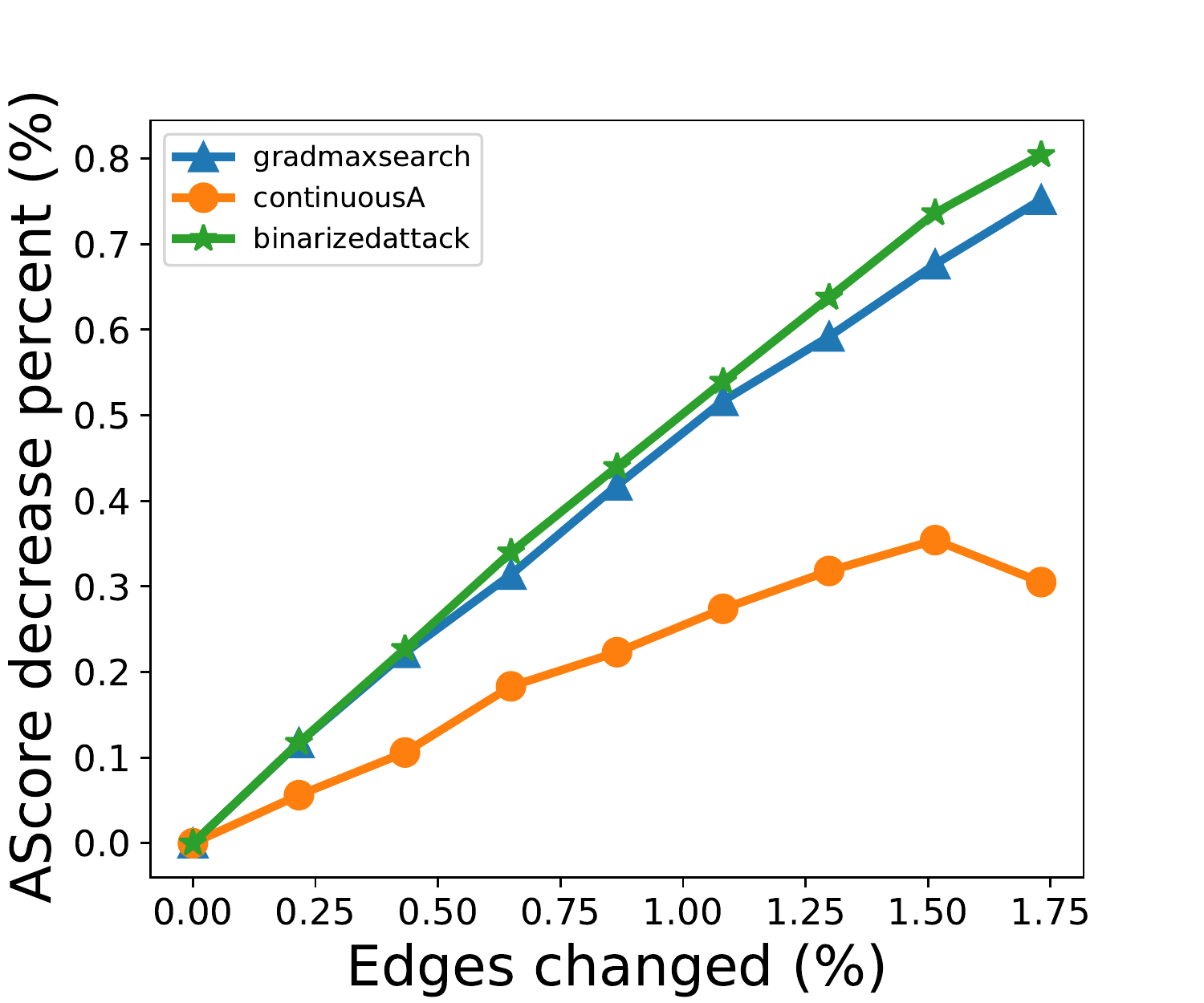}
		\caption{\textbf{Bitcoin-Alpha}-10}
	\end{subfigure}
	\hfill
	\begin{subfigure}[b]{0.234\textwidth}
		\centering
		\includegraphics[width=\textwidth,height=3cm]{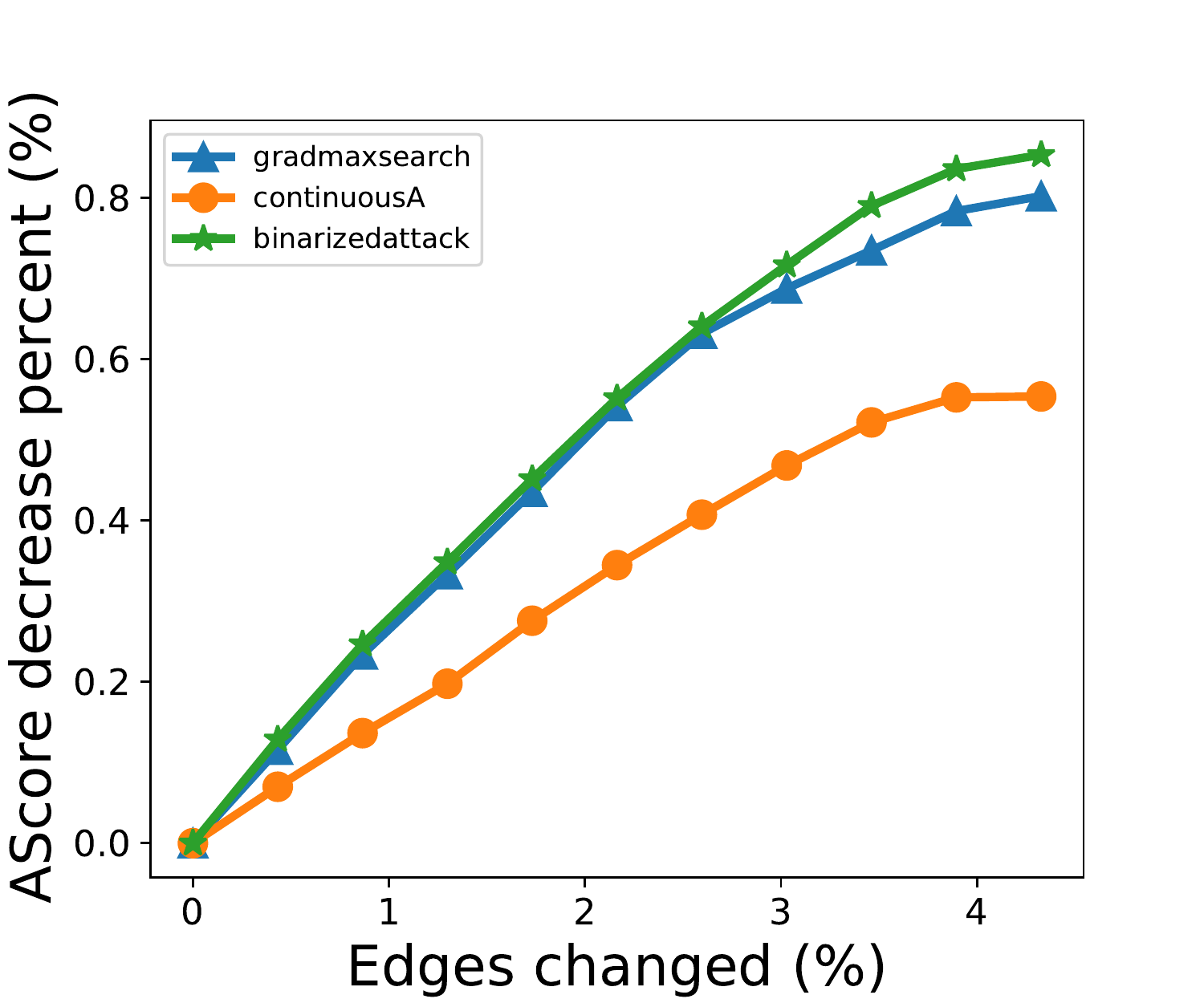}
		\caption{\textbf{Bitcoin-Alpha}-30}
	\end{subfigure}
	\hfill
	\begin{subfigure}[b]{0.234\textwidth}
		\centering
		\includegraphics[width=\textwidth,height=3cm]{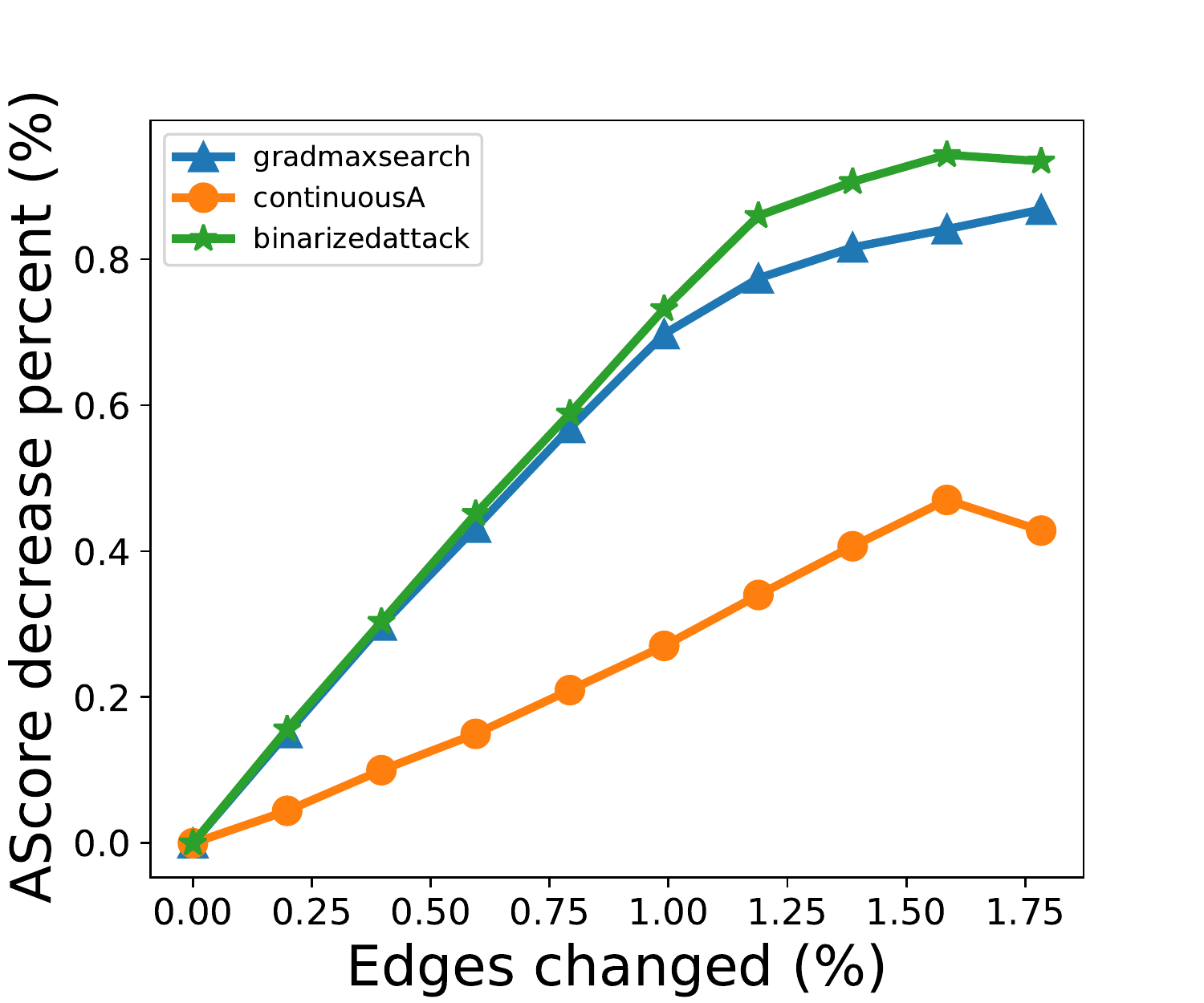}
		\caption{\textbf{Wikivote}-10}
	\end{subfigure}
	\hfill
	\begin{subfigure}[b]{0.234\textwidth}
		\centering
		\includegraphics[width=\textwidth,height=3cm]{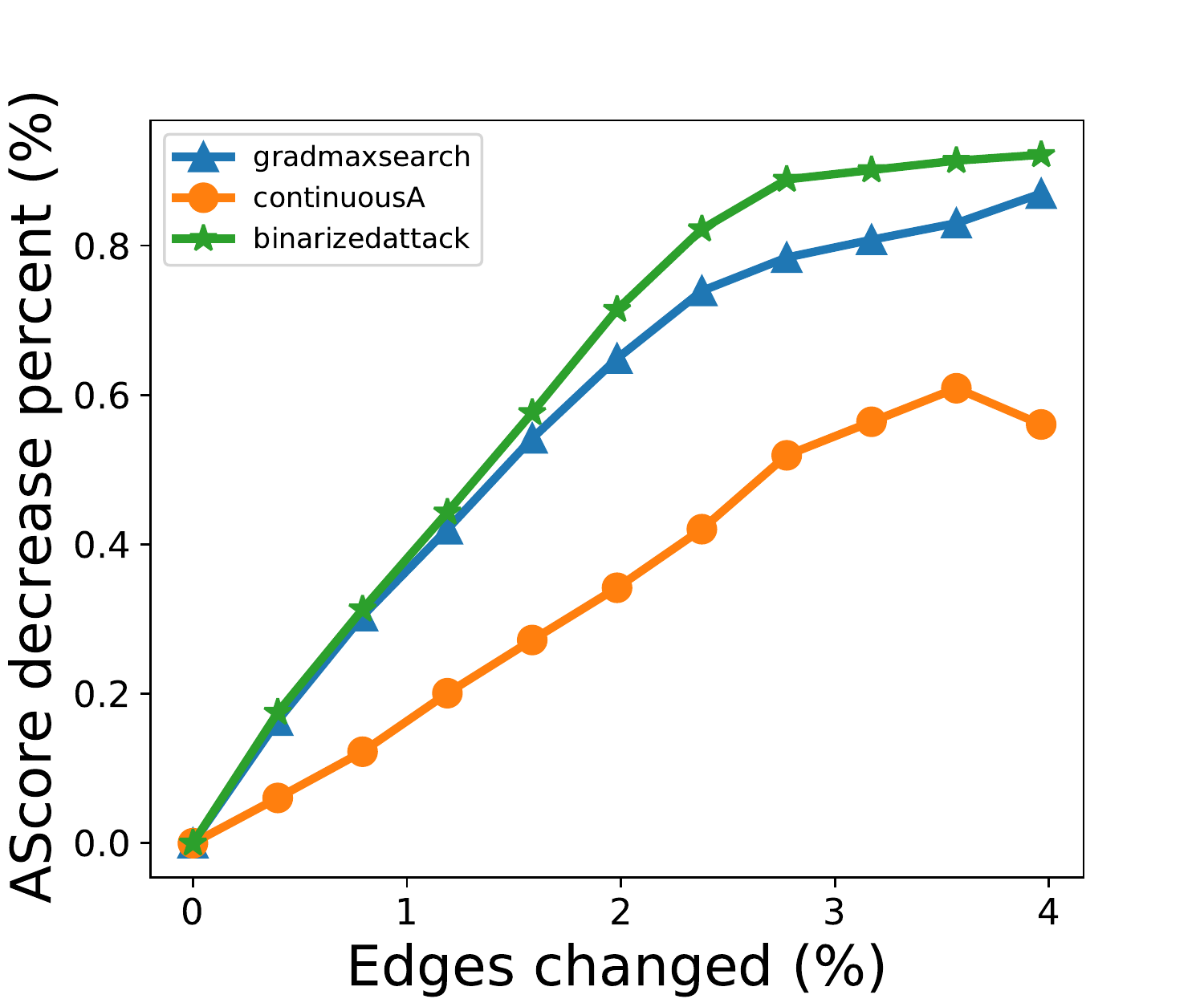}
		\caption{\textbf{Wikivote}-30}
		\label{fig-wiki-30}
	\end{subfigure}
	\caption{Changes in $\mathsf{AScores}$ for \textbf{ER} graph, \textbf{BA} graph, \textbf{Bitcoin-Alpha}, \textbf{Blogcatalog} and \textbf{Wikivote}}
	\label{fig-main-exp}
\end{figure}

\subsubsection{Effectiveness of attack} Our main focus is to investigate
whether the proposed attack methods can effectively decrease $\mathsf{AScores}$ of the target nodes. We use the \textbf{Decreasing
Percentage of Anomaly Score} denoted as $\tau_{as}$ as the evaluation metric. Specifically, let $S_{\mathcal{T}}^0$ and $S_{\mathcal{T}}^B$ be the sum of $\mathsf{AScores}$ of the target nodes before attack and after an
attack with a budget $B$, respectively. Then, $\tau_{as}$ is defined as
$\tau_{as} = (S_{\mathcal{T}}^0 - S_{\mathcal{T}}^B) / S_{\mathcal{T}}^0$.

Fig.~\ref{fig-main-exp} presents the $\tau_{as}$ of three attack methods,
$\mathsf{BinarizedAttack}$, $\mathsf{GradMaxSearch}$, and $\mathsf{ContinuousA}$, with varying attack power. In particular, the attack power is measured as a percentage $\frac{B}{|E|}$, where $|E|$ is the number of edges in the clean graph. We note that in all the cases, the attacker
has modified very limited edges in the graph: less than $2\%$
when $|V | = 10$ and less than $5\%$ when $|V | = 30$. On average,
for each target node, the number of modified edges is around 4 for \textbf{Bitcoin-Alpha}, 5 for \textbf{Blogcatalog} and 9 for \textbf{Wikivote}.

We can observe from Fig.~\ref{fig-main-exp} that both $\mathsf{BinarizedAttack}$
and $\mathsf{GradMaxSearch}$ can significantly (up to $90\%$) decrease $\mathsf{AScores}$ with very limited power while
$\mathsf{ContinuousA}$ is not effective in several cases. This demonstrated the unpredictable feature of $\mathsf{ContinuousA}$ when converting continuous solutions to discrete ones. Our main method $\mathsf{BinarizedAttack}$ consistently achieves the best performance in
all cases. We note that the margin between $\mathsf{BinarizedAttack}$ and $\mathsf{GradMaxSearch}$ is significant (with a $> 10\%$ performance
improvement) when the attack power is high (although the two lines look close in the figures). For example, in Fig.~\ref{fig-wiki-30},
when the attack power is $2.77\%$, $\mathsf{BinarizedAttack}$ outperforms
$\mathsf{GradMaxSearch}$ by $13.32\%$. One intriguing observation is
that the gap between $\mathsf{BinarizedAttack}$ and $\mathsf{GradMaxSearch}$
becomes larger when we increase the attacker’s budget. The
reason is that $\mathsf{GradMaxSearch}$ is a greedy strategy in nature
and should performs well when modifying a few edges;
meanwhile, it is also myopic when the budget is large.

\begin{figure}
	\centering
	\begin{subfigure}[b]{0.24\textwidth}
		\centering
		\includegraphics[width=\textwidth]{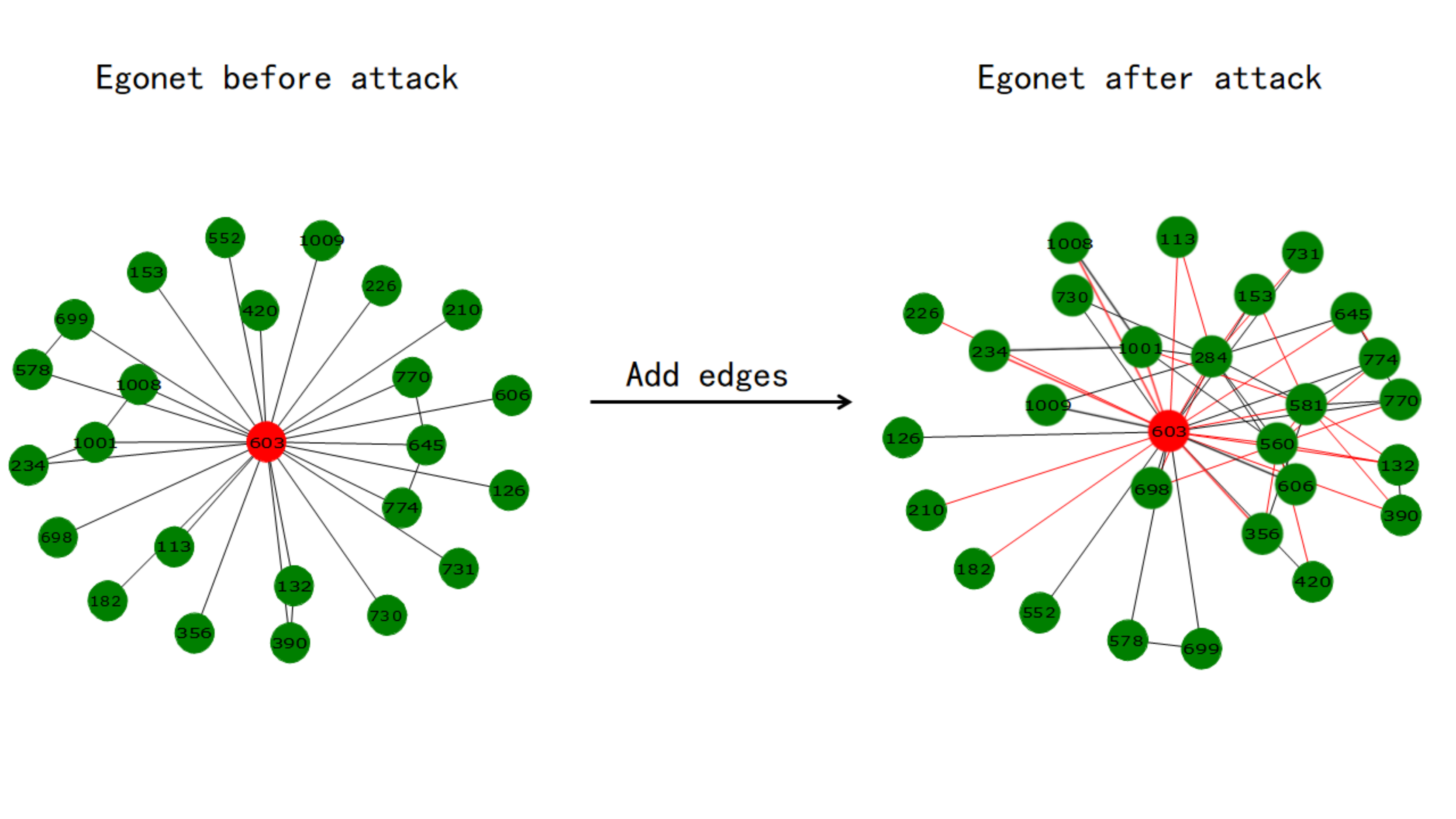}
		\caption{Case 1: Add edges}
	\end{subfigure}
	\hfill
	\begin{subfigure}[b]{0.24\textwidth}
		\centering
		\includegraphics[width=\textwidth]{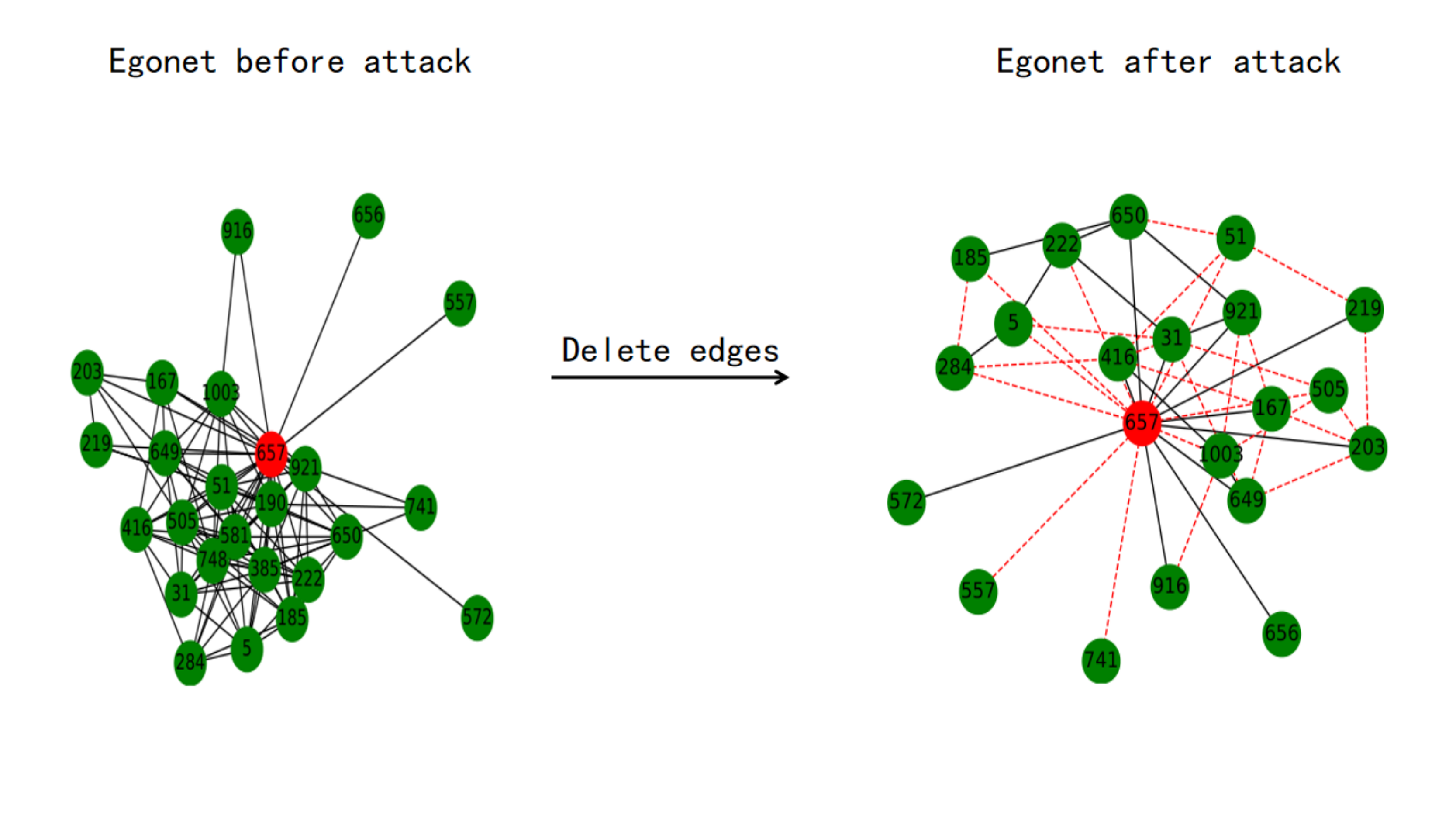}
		\caption{Case 2: Delete edges}
	\end{subfigure}
	\hfill
	\begin{subfigure}[b]{0.24\textwidth}
		\centering
		\includegraphics[width=\textwidth]{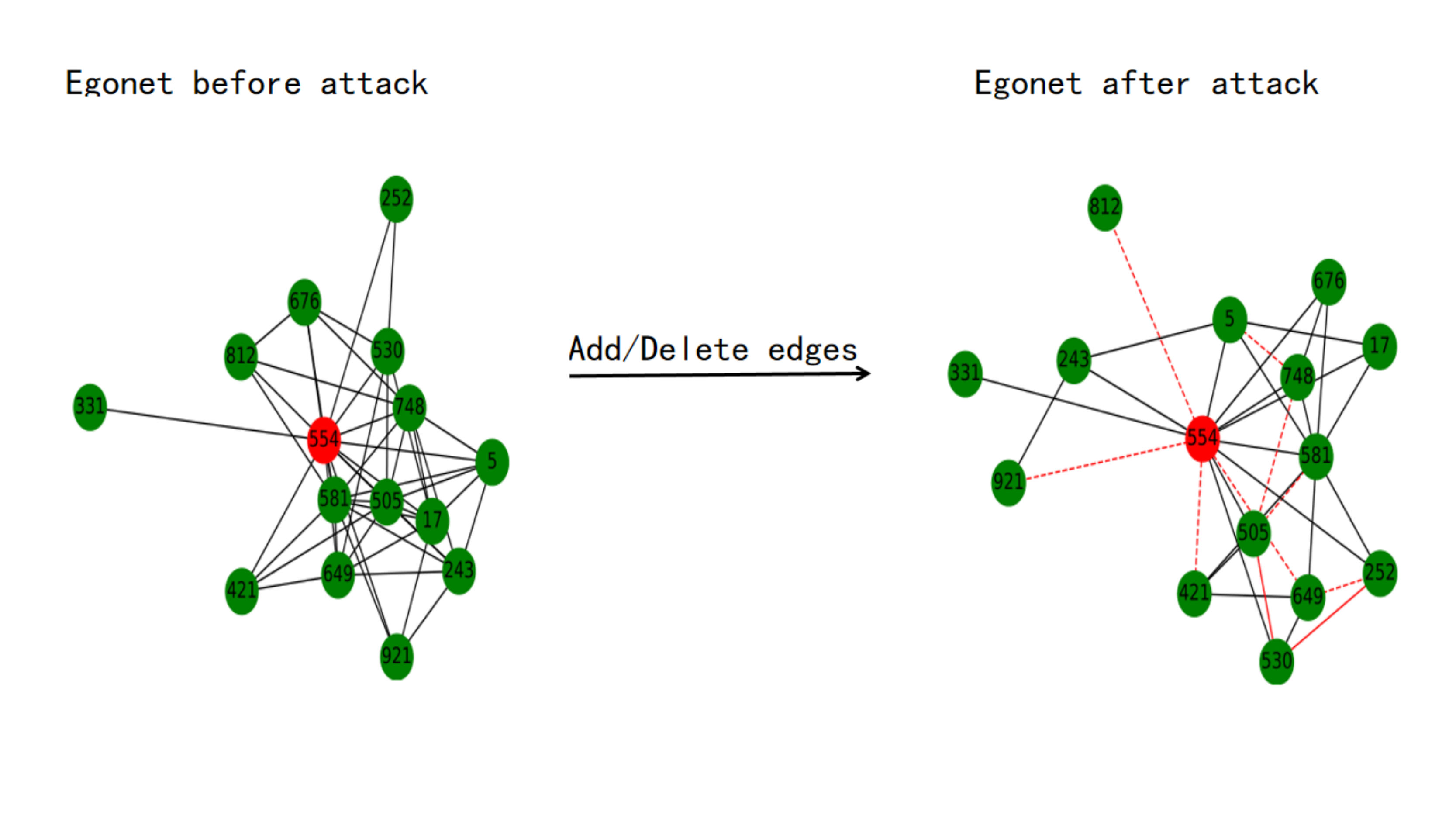}
		\caption{Case 3: Add/Delete edges}
	\end{subfigure}
	\caption{Attacker implemented $\mathsf{BinarizedAttack}$ method on the three target nodes $v_{603}$, $v_{657}$ and $v_{554}$. In the case 1, the attacker add edges to the clean graph, the $\mathsf{AScore}$ decreases from 6.05 to 0.69; in the case 2, the attack delete the edges to the clean graph, the $\mathsf{AScore}$ decreases from 8.4 to 0.29; in the case 3, the attacker add and delete edges simultaneously to decrease the anomaly node $v_{554}$'s $\mathsf{AScore}$ from 5.34 to 0.42. We observe that in these three cases $\mathsf{BinarizedAttack}$ indeed force the \textit{near-star} and \textit{near-clique} Egonets to become 'normal' ones, i.e., the edges density inside the Egonets is at a rational level}
	\label{fig-visual}
\end{figure}

We further show in Fig.~\ref{fig-visual} how $\mathsf{BinarizedAttack}$ will actually modify the structure of a real-world graph (\textbf{Wikivote} as the example). We
demonstrated three cases, where the attacker deletes edges
only, adds edges only, and deletes and adds edges simultaneously. It shows that $\mathsf{BinarizedAttack}$ will \textit{indeed break the
anomalous structural patterns (i.e., near-star and near-clique)
in the graph}.



\subsubsection{Preferences of attack} 
\begin{figure}
	\centering
	\begin{subfigure}[b]{0.15\textwidth}
		\centering
		\includegraphics[width=\textwidth]{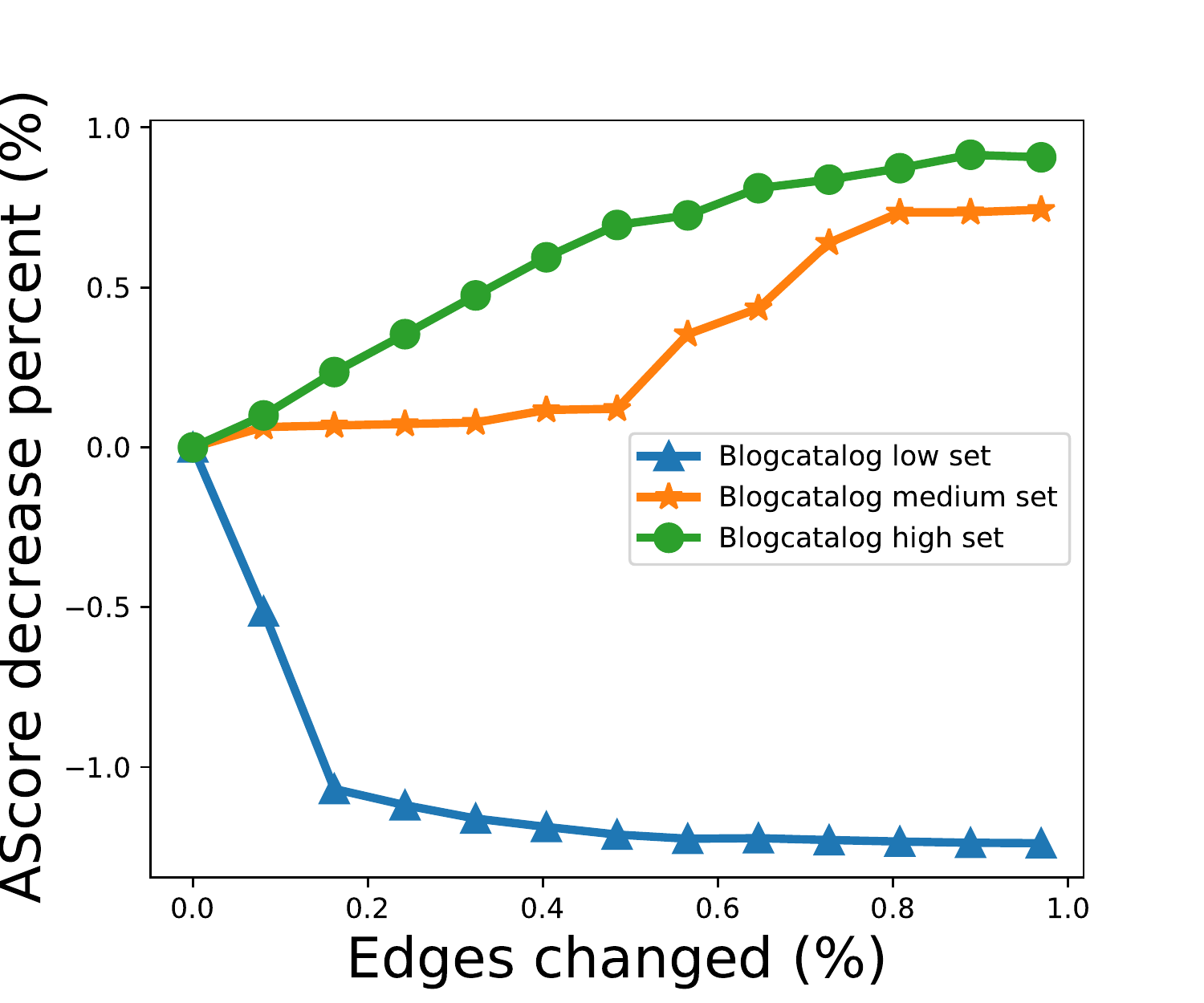}
		\caption{$\mathsf{AScores}$ decreasing percentage}
	\end{subfigure}
	\hfill
	\begin{subfigure}[b]{0.15\textwidth}
		\centering
		\includegraphics[width=\textwidth]{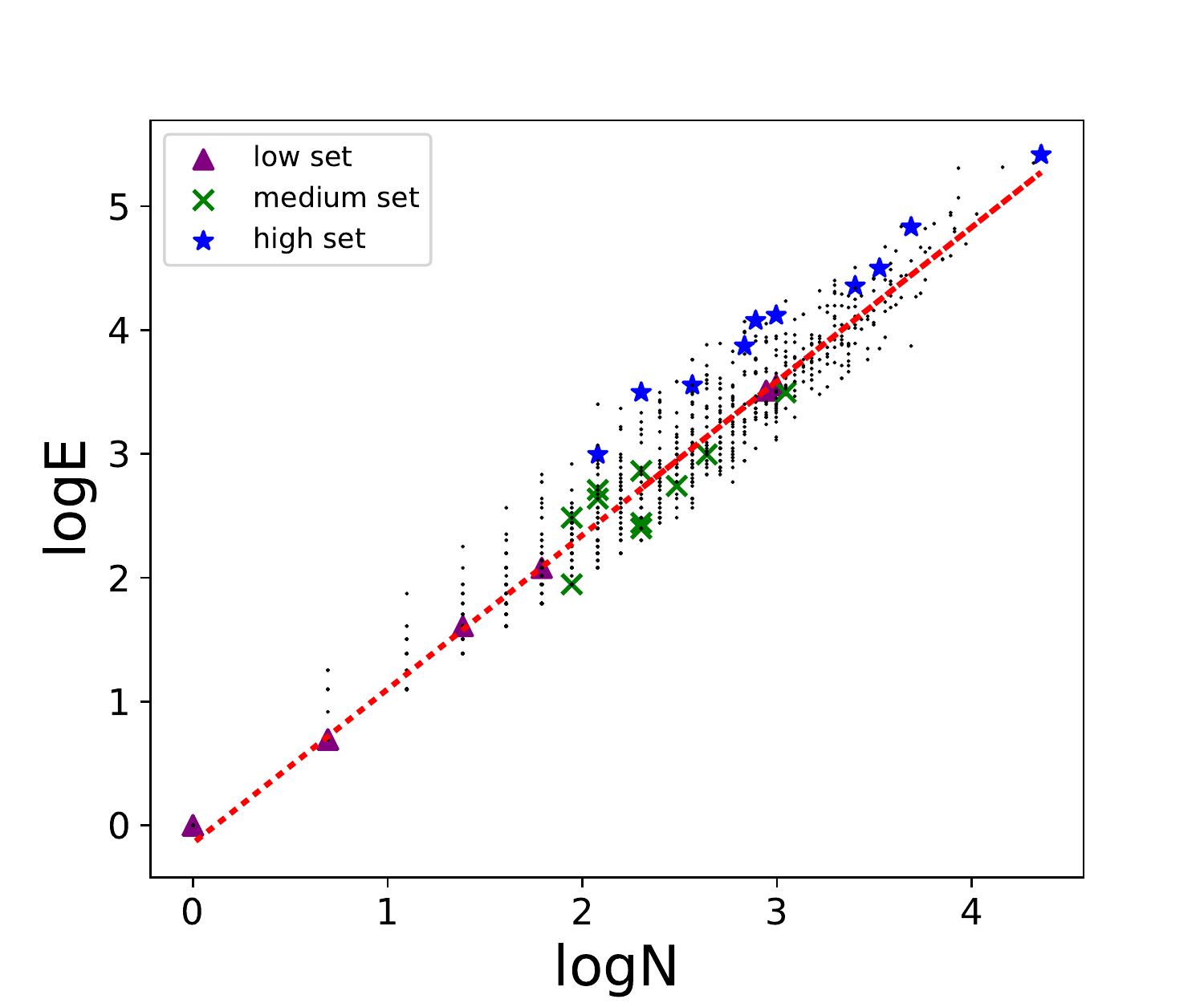}
		\caption{Linear regression with $B=0$}
	\end{subfigure}
	\hfill
	\begin{subfigure}[b]{0.15\textwidth}
		\centering
		\includegraphics[width=\textwidth]{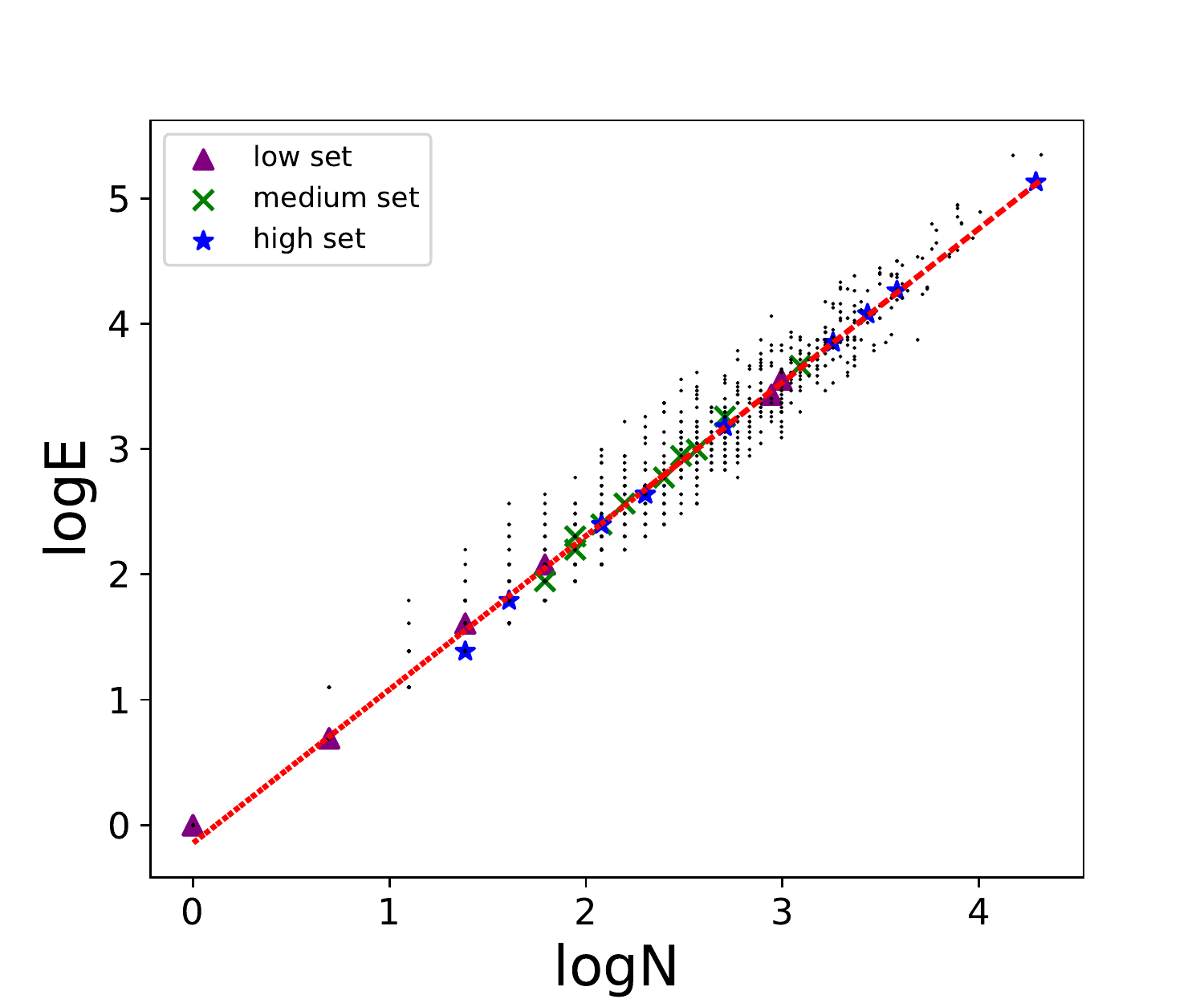}
		\caption{Linear regression with $B=60$}
	\end{subfigure}
	\caption{Structural attack on \textbf{Blogcatalog} with different target set. (a) is the decreasing percentage of $\mathsf{AScores}$ for different cases; (b) is linear regression for the clean graph; (c) is linear regression for the poisoned graph.}
	\label{fig-split}
\end{figure}

In the previous experiment evaluation, we select a set of target nodes that have relatively high $\mathsf{AScores}$ and measure the attack performance by the
decreasing percentage of the score sum as a whole. It is also
important to investigate the different attack effects on individual targets. We thus group the nodes according to their initial $\mathsf{AScores}$ in the clean graph. Specifically, we assign the 10\% percentile and 90\% percentile ($q_{1}$ and $q_{2}$) 
for $\mathsf{AScores}$ histogram as thresholds. Using the thresholds $q_1$ and $q_2$, we split all the nodes into
three groups: low level, medium level, and high level, based
on their $\mathsf{AScores}$. We randomly pick out $10$ nodes from
each group and treat the $30$ nodes as the target set $\mathcal{T}$. We report
the decreasing percentage $\tau_{as}$ as under $\mathsf{BinarizedAttack}$ for each group separately in Fig.~\ref{fig-split}. We observe that the changes of the distance between high-level nodes and the regression lines with clean graph and poisoned graph is larger than that of low-level and medium-level nodes. This phenomenon supports the saying that $\mathsf{BinarizedAttack}$ tends to exert more influence on high-level anomaly nodes.

\subsubsection{Side effect of attack}
$\mathsf{BinarizedAttack}$ is a poisoning targeted attack that aims to mislead the predictions of a \textit{small set of} target nodes. A desirable feature is that the attack would not significantly change the data distribution to the extent that the poisoned data would look abnormal for a defender. In other words, an attacker wants to make the attack \textit{unnoticeable}.

In our context, $\mathsf{BinarizedAttack}$ modifies the features $(\mathbf{N}_i, \mathbf{E}_i)$ of the ego-net centered at node $i$. By design, $\mathsf{BinarizedAttack}$ will significantly modify the features of targeted nodes; however, an inevitable \textit{side effect} is that $\mathsf{BinarizedAttack}$ would also change the features of the rest of the nodes. In the worst case, the shift of the feature distributions caused by an attack is so large that the attacked graph would appear abnormal and be rejected by the defender. Thus, ideally, we would expect the side effect of $\mathsf{BinarizedAttack}$ is reasonably small to achieve an unnoticeable attack. 

We thus experiment to investigate the distribution shift of the features $(\mathbf{N}, \mathbf{E})$ before and after $\mathsf{BinarizedAttack}$.
Specifically, we use the permutation test \cite{permutationtest}, which is a non-parametric test to check whether two different groups follow the same distribution. However, it will be NP-hard if we consider all kinds of perturbations in the concatenation of two group data, we instead use the Monte Carlo method to approximate the \textit{p}-value, which is computed as:

\begin{align}
\label{eqn-pvalue}
p(t\geq t_{0})=\frac{1}{M}\sum_{j=1}^{M}I(t_{j}\geq t_{0}),
\end{align}
where $t_{0}$ is the observed value of test statistic and $t$ is the statistic calculated from the re-samples, i.e., $t(x_{1}^{\prime},x_{2}^{\prime},...,x_{n}^{\prime},y_{1}^{\prime},y_{2}^{\prime},...,y_{n}^{\prime})=|\bar{x^{\prime}}-\bar{y^{\prime}}|,$ $M$ is the Monte Carlo sampling times (in the experiment we set $M=100000$). $x$ and $y$ can be either $\mathbf{N}^{clean}$ and $\mathbf{N}^{poisoned}$ or $\mathbf{E}^{clean}$ and $\mathbf{E}^{poisoned}$. $x^{\prime}$ and $y^{\prime}$ are re-samples of $\mathsf{Concat}[\mathbf{N}^{clean}||\mathbf{N}^{poisoned}]$ or $\mathsf{Concat}[\mathbf{E}^{clean}||\mathbf{E}^{poisoned}]$.

We report the $p$-value of ego-features $\mathbf{N}$ and $\mathbf{E}$ on three real datasets: \textbf{Bitcoin-Alpha}, \textbf{Blogcatalog} and \textbf{Wikivote}. The results are shown in Tab.~\ref{tab-unnotice}, where we consider ego-features of the poisoned graph with maximum perturbations and $|V|=30$.

\begin{table}[h]
	\centering
	\caption{$p$-values for ego-features.}
	\label{tab-unnotice}
	\begin{tabular}{|c|cc|cc|cc|}
		\hline
        \multirow{2}*{\diagbox{$t$}{$p$}{features}}&\multicolumn{2}{c|}{\textbf{Bitcoin-Alpha}}&\multicolumn{2}{c|}{\textbf{Blogcatalog}}&\multicolumn{2}{c|}{\textbf{Wikivote}}\\
		&$\mathbf{N}$&$\mathbf{E}$&$\mathbf{N}$&$\mathbf{E}$&$\mathbf{N}$&$\mathbf{E}$\\
		\hline
		1&0.72&0.04&0.72&0.12&0.56&0.02\\
		\hline
		2&0.66&0.03&0.65&0.06&0.58&0.02\\
		\hline
		3&0.75&0.04&0.67&0.06&0.56&0.01\\
		\hline
		4&0.72&0.04&0.71&0.12&0.57&0.02\\
		\hline
		5&0.71&0.03&0.7&0.14&0.56&0.005\\
		\hline
	\end{tabular}
\end{table}

\begin{figure}[h]
	\centering
	\begin{subfigure}[b]{0.234\textwidth}
		\centering
		\includegraphics[width=\textwidth,height=3cm]{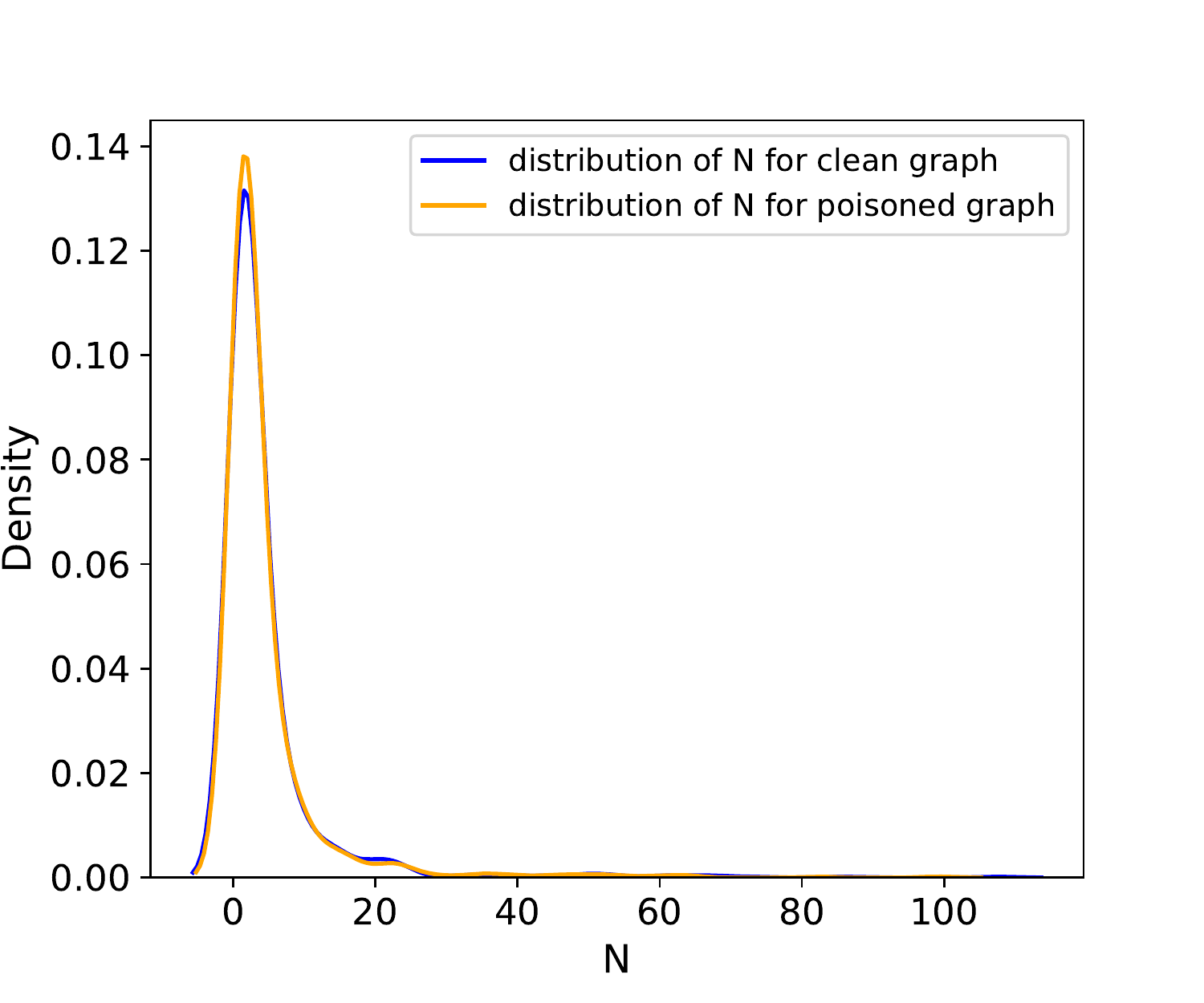}
		\caption{\textbf{Bitcoin-Alpha N}}
	\end{subfigure}
	\hfill
	\begin{subfigure}[b]{0.234\textwidth}
		\centering
		\includegraphics[width=\textwidth,height=3cm]{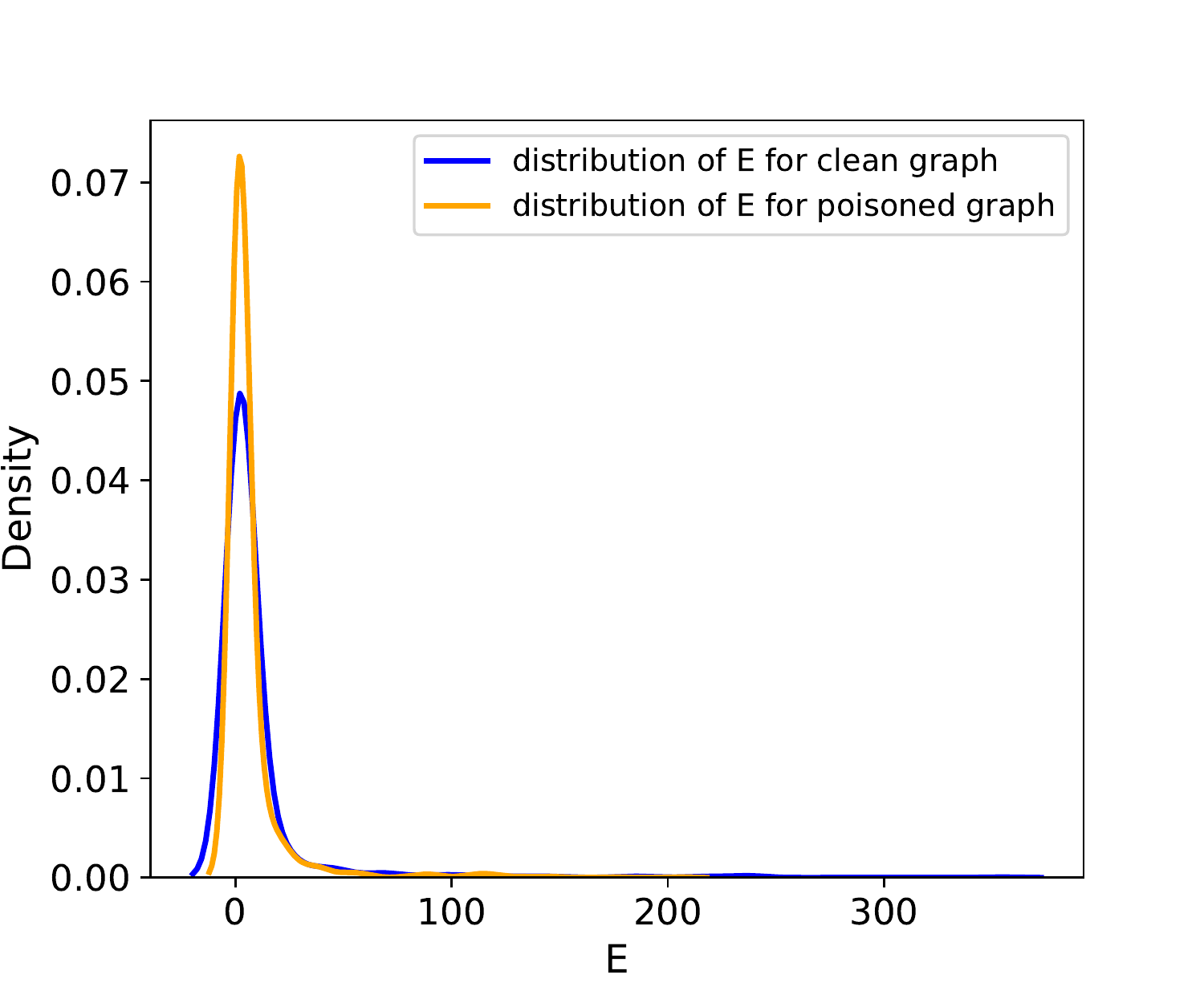}
		\caption{\textbf{Bitcoin-Alpha E}}
	\end{subfigure}
	\caption{The probability density function of ego-features $\textbf{N}$ and $\textbf{E}$ for clean and poisoned graphs of \textbf{Bitcoin-Alpha} }
\label{fig-kde}
\end{figure}

From Tab.~\ref{tab-unnotice} we notice that under 99\% significant level we cannot reject the null hypothesis that $\mathbf{N}^{clean}$ and $\mathbf{N}^{poisoned}$ follows the same distribution, that is, we can draw the conclusion that $\mathsf{BinarizedAttack}$ does not manipulate the distribution of $\mathbf{N}$. For ego-feature $\mathbf{E}$, in most cases, we cannot reject the null hypothesis. However, in experiment 5 for \textbf{Wikivote} the $p$-value is less than 1\%, so we reject the null hypothesis, that is, in \textbf{Wikivote}   experiment 5 $\mathsf{BinarizedAttack}$ manipulate the distribution of $\textbf{E}$ under the significant level 99\%. The probability density function of $\textbf{N}^{clean}$ and $\textbf{N}^{poisoned}$ and $\textbf{E}^{clean}$ and $\textbf{E}^{poisoned}$ are presented in Fig.~\ref{fig-kde} (we take \textbf{Bitcoin-Alpha} as an example).

\subsection{Attack Transferability}
\label{sec-transfer-GAL}
We follow the procedure in Section~\ref{sec-transfer-method} to conduct transfer attacks against GAL and ReFeX on \textbf{Bitcoin-Alpha} and \textbf{Wikivote}. We use the configurations in \cite{10.1145/3340531.3411979} and \cite{ReFeX} to implement GAL and ReFeX, respectively. We measure the attack effects on classification results as well as node embeddings.
\subsubsection{Classification results}
$\mathsf{BinarizedAttack}$ is a targeted attack. To evaluate the effects more comprehensively, we measure both the global and targeted classification accuracies of GAL and ReFeX under attack. Specifically, for global classification accuracies, we compare the AUC and F1 scores on all test nodes before and after attack. For targeted classification accuracies, we use the changes in \textit{soft labels} \cite{softlabel} of the targeted nodes as the metric. A \textit{soft label} is a score (probability) of an entity which belong to one class in a classifier.
Specifically, in the representation learning literature, the sum of \textit{soft labels} for target nodes is defined as: $\sum_{i=1}^{K}P_{i}(Y_{i}=1|\theta)=\sum_{i=1}^{K}f_{\theta}(X_{emb})$, $f_\theta (\cdot)$ is the classifier MLP with parameter $\theta$,
$X_{emb}$ is the node embeddings learned by GAL or ReFeX. Referring to \cite{softlabel}, \textit{soft labels} have high entropy which provide much more information than the hard labels, thus can capture the quality of embeddings in this scenario more precisely. Let $\mathcal{SL}_0$ and $\mathcal{SL}_B$ denote the sum of \textit{soft labels} before and after attack with budget $B$. 
We use the \textbf{decreasing percentage of soft labels} as the metric, defined as $\delta_B = (\mathcal{SL}_0 - \mathcal{SL}_B)/\mathcal{SL}_0$.

The classification results of GAL and ReFeX are presented in Table~\ref{tab-GAL} and Table~\ref{tab-ReFeX}. We can observe that $\mathsf{BinarizedAttack}$ have limited effect on the global classification accuracies of both GAL and ReFeX (refer to the AUC and F1 columns). However, $\mathsf{BinarizedAttack}$ can significantly decrease the \textit{soft labels} (around $25\%$) with very limited budgets. We emphasize that this observation actually reflects a desirable feature of $\mathsf{BinarizedAttack}$ as a targeted attack. That is, $\mathsf{BinarizedAttack}$ would not severely downgrade the overall performance of the system, which avoids attracting attention from system defenders; however, it can significantly mislead the system's predictions on the targets.

\begin{table}[h]
	\centering
	\caption{Classification accuracy for the GAL embeddings provided for the anomaly classification task.}
	\label{tab-GAL}
	\resizebox{0.9\columnwidth}{!}{%
	\begin{tabular}{|c|ccc|ccc|}
		\hline
		& \multicolumn{3}{c|}{\textbf{Bitcoin-Alpha}} & \multicolumn{3}{c|}{\textbf{Wikivote}}\\
		Edge changed (\%) & AUC & F1 & $\delta_B$ (\%)& AUC & F1 & $\delta_B$ (\%)\\
		\hline
		\hline
		  0& 0.72&0.85&    0&0.68&0.77&0\\
		0.2& 0.71&0.84&10.87&0.67&0.74&19.49\\
		0.4& 0.71&0.84&14.20&0.65&0.72&19.43\\
		0.6& 0.68&0.82&14.46&0.68&0.76&23.29\\
		0.8& 0.69&0.83&16.30&0.65&0.72&24.55\\
		1.0& 0.67&0.82&20.00&0.65&0.76&26.61\\
		1.2& 0.68&0.83&18.42&0.59&0.69&28.02\\
		1.4& 0.68&0.82&18.76&0.65&0.72&26.00\\
		1.6& 0.65&0.81&23.16&0.66&0.74&23.97\\
		1.8& 0.66&0.82&21.60&0.66&0.75&26.45\\
		2.0& 0.65&0.81&25.69& 0.6&0.71&25.25\\
		\hline
	\end{tabular}
}
\end{table}

\begin{table}[h]
	\centering
		\caption{Classification accuracy for the ReFeX embeddings provided for the anomaly classification task}
	\label{tab-ReFeX}
		\resizebox{0.9\columnwidth}{!}{%
	\begin{tabular}{|cccc|cccc|}
		\hline
		\multicolumn{4}{|c|}{\textbf{Bitcoin-Alpha}} & \multicolumn{4}{c|}{\textbf{Wikivote}}\\
		B & AUC & F1 & $\delta_B$ (\%) & B & AUC & F1 & $\delta_B$ (\%)\\
		\hline
		\hline
		0& 0.79& 0.9 &    0&  0&0.84&0.92&0\\
		5& 0.76&0.89 &15.15& 10&0.81&0.92&16.52\\
		10& 0.77&0.89&15.50& 20&0.79&0.9 &20.55\\
		15& 0.76& 0.9&19.81& 30&0.81&0.91&26.00\\
		20& 0.74&0.89&21.42& 40&0.72&0.88&36.31\\
		25& 0.72&0.88&25.30& 50& 0.7&0.87&41.40\\
		30& 0.73&0.89&24.91& 60&0.68&0.88&49.52\\
		35& 0.72&0.88&31.78& 70& 0.7&0.88&54.71\\
		40& 0.73&0.89&31.05& 80&0.66&0.87&56.16\\
		45& 0.75&0.88&31.34& 90&0.64&0.86&56.02\\
		50& 0.72&0.88&33.27&100&0.66&0.87&56.37\\
		\hline
	\end{tabular}
}
\end{table}



\subsubsection{Node embeddings}
As both GAL and ReFeX are representation-learning-based methods, it is important to see the qualities of the node embeddings under attack. We thus further provide the visualization of node embeddings before and after attack to highlight the differences. To this end, we use a dimension reduction method t-SNE \cite{tsne} to visualize the penultimate hidden features of the classifier MLP of the testing nodes in 2-D space. Referring to \cite{DBLP:journals/corr/abs-1808-05385}, the penultimate hidden features of the MLP will form the linear separable space in 2-D space after dimension reduction (PCA \cite{pca} or t-SNE \cite{tsne}).

\begin{figure}[h]
	\centering
	\begin{subfigure}[b]{0.234\textwidth}
		\centering
		\includegraphics[width=\textwidth,height=3cm]{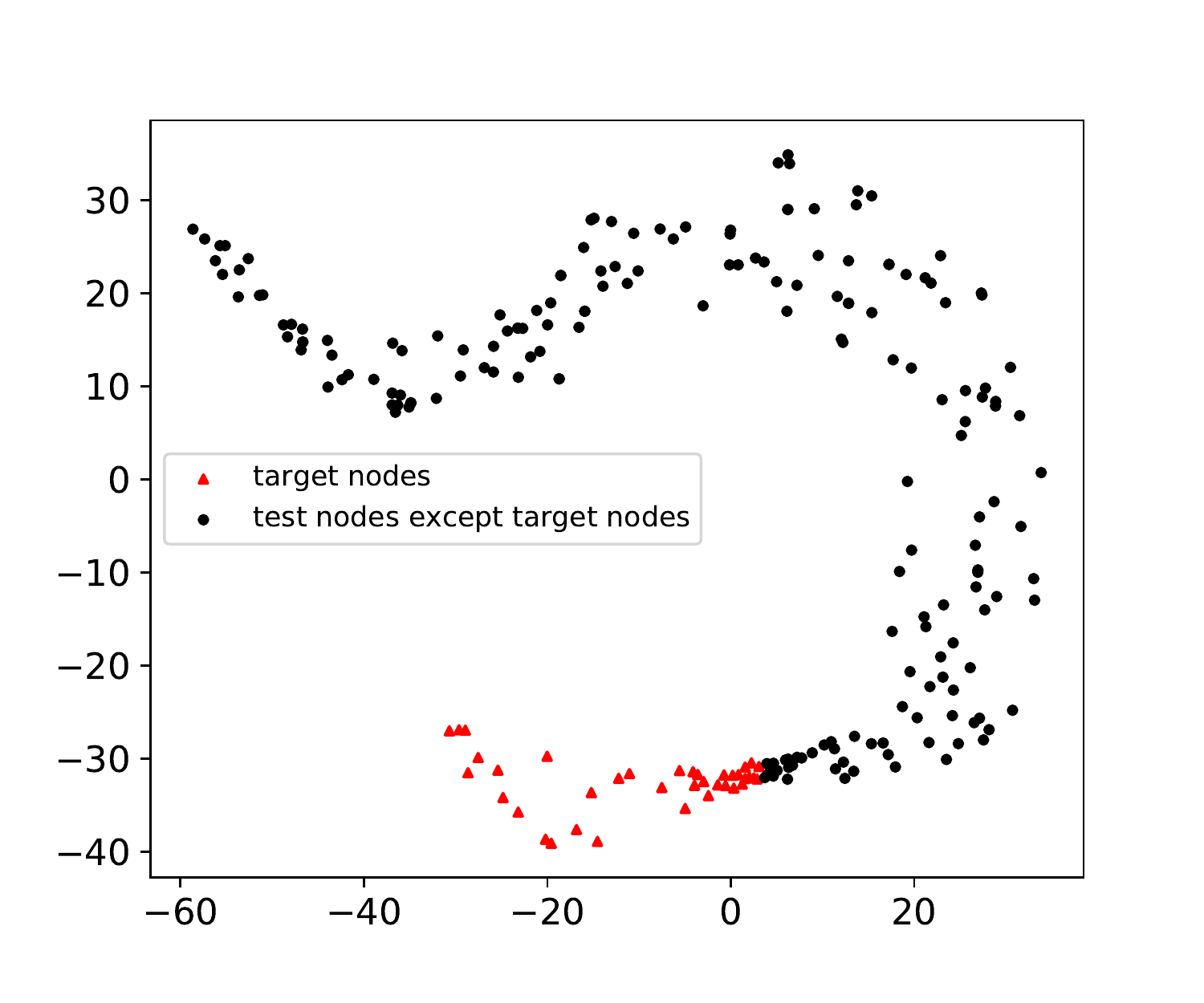}
		\caption{\textbf{Bitcoin-Alpha} clean graph }
	\end{subfigure}
	\hfill
	\begin{subfigure}[b]{0.234\textwidth}
		\centering
		\includegraphics[width=\textwidth,height=3cm]{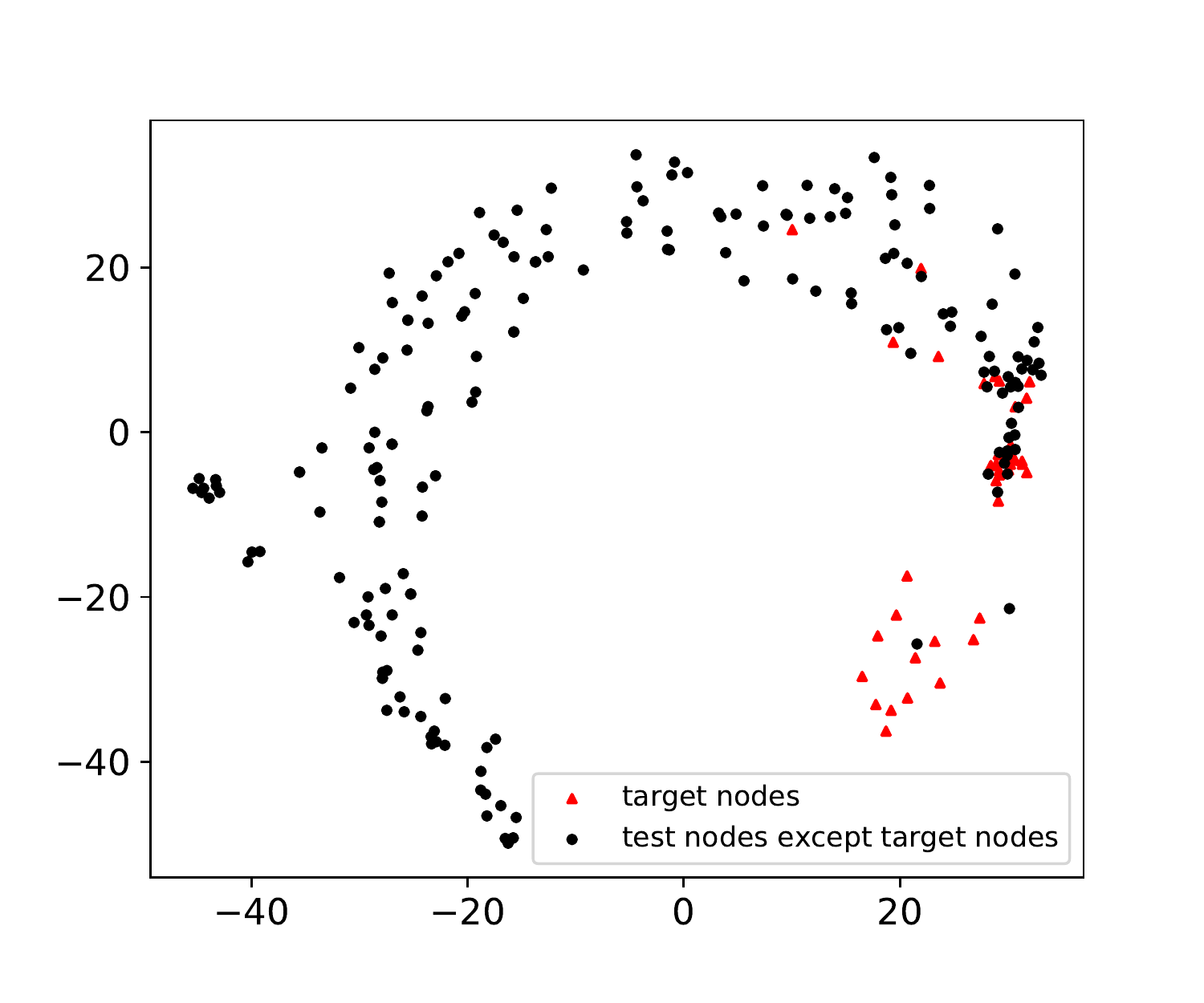}
		\caption{\textbf{Bitcoin-Alpha} with $B=50$}
	\end{subfigure}
    \hfill
	\begin{subfigure}[b]{0.234\textwidth}
	\centering
	\includegraphics[width=\textwidth,height=3cm]{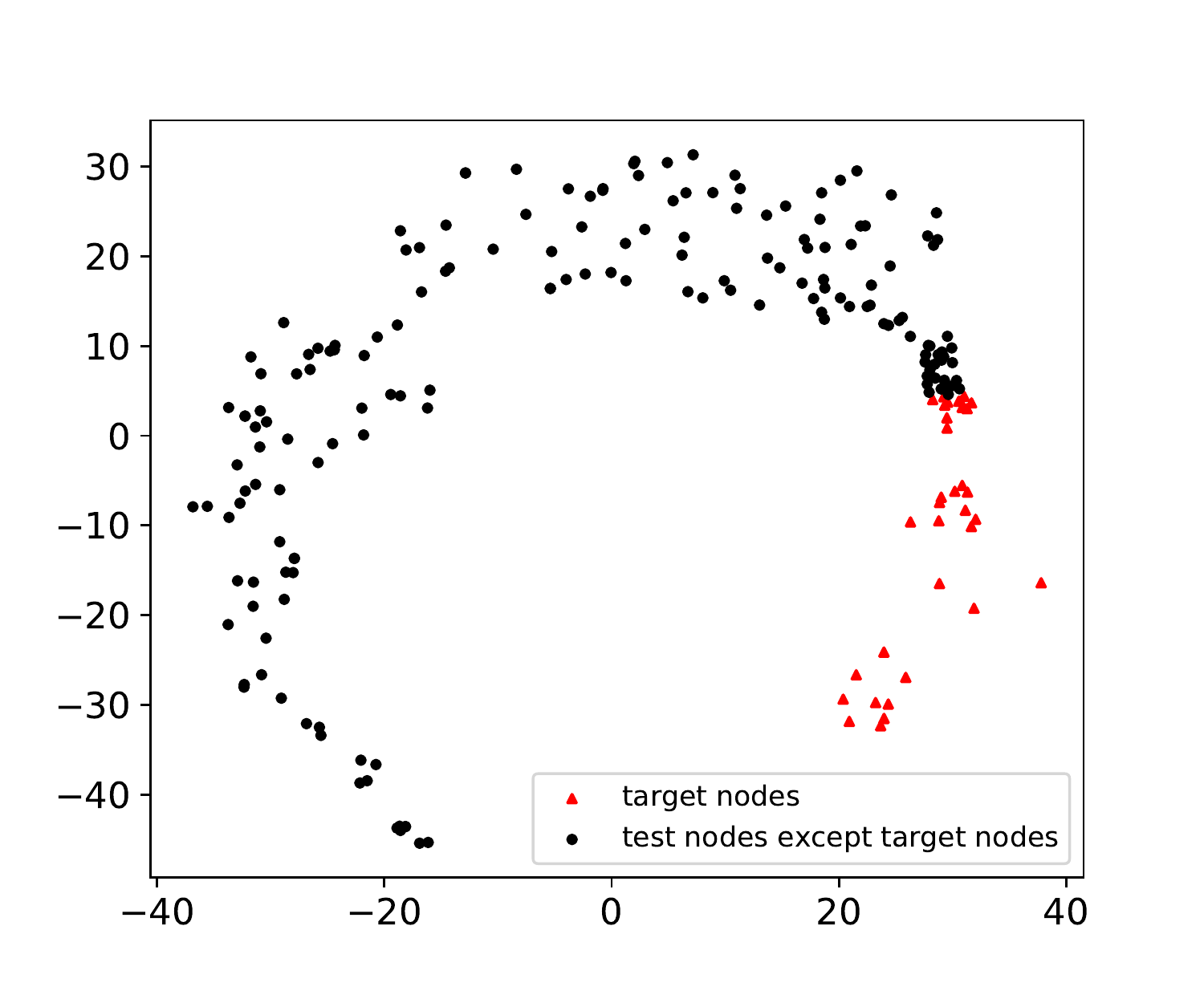}
	\caption{\textbf{Wikivote} clean graph}
     \end{subfigure}
\hfill
\begin{subfigure}[b]{0.234\textwidth}
	\centering
	\includegraphics[width=\textwidth,height=3cm]{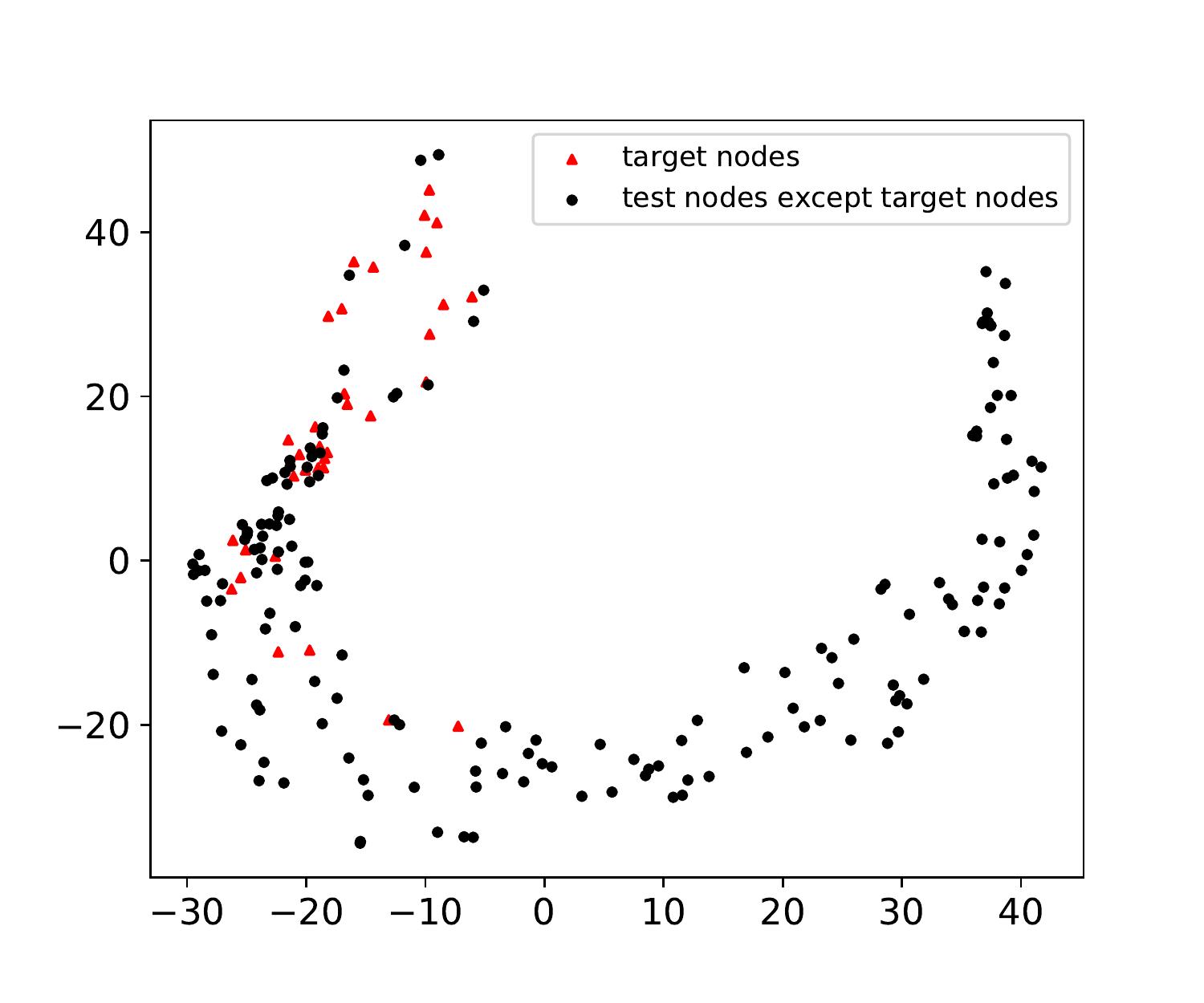}
	\caption{\textbf{Wikivote} with $B=100$}
\end{subfigure}
	\caption{Scatterplot of GAL embeddings of penultimate hidden features for clean graph and poisoned graph of \textbf{Bitcoin-Alpha} and \textbf{Wikivote} in 2-D space}
\label{fig-emb-gal}
\end{figure}

\begin{figure}[h]
	\centering
	\begin{subfigure}[b]{0.234\textwidth}
		\centering
		\includegraphics[width=\textwidth,height=3cm]{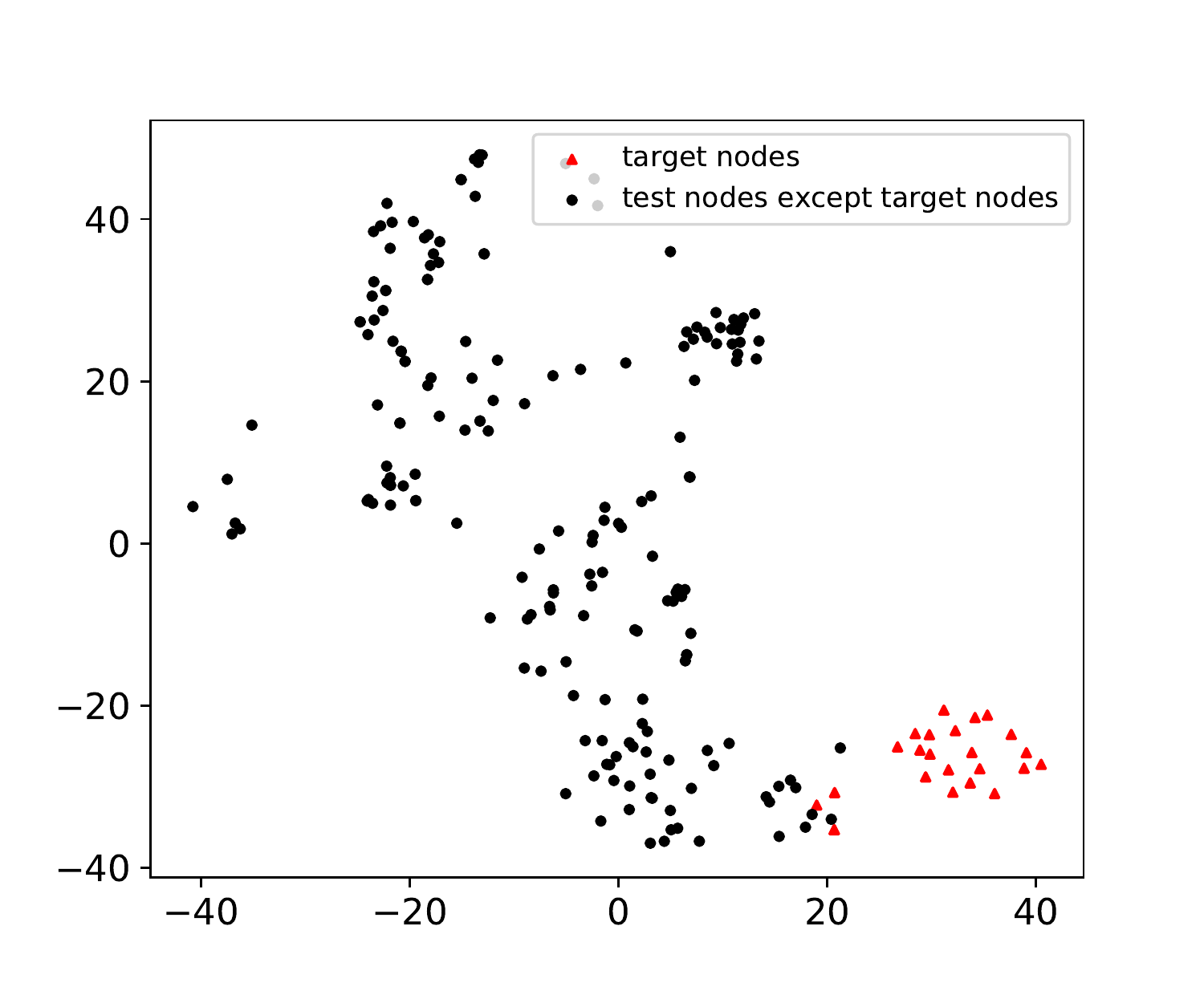}
		\caption{\textbf{Bitcoin-Alpha} clean graph}
	\end{subfigure}
	\hfill
	\begin{subfigure}[b]{0.234\textwidth}
		\centering
		\includegraphics[width=\textwidth,height=3cm]{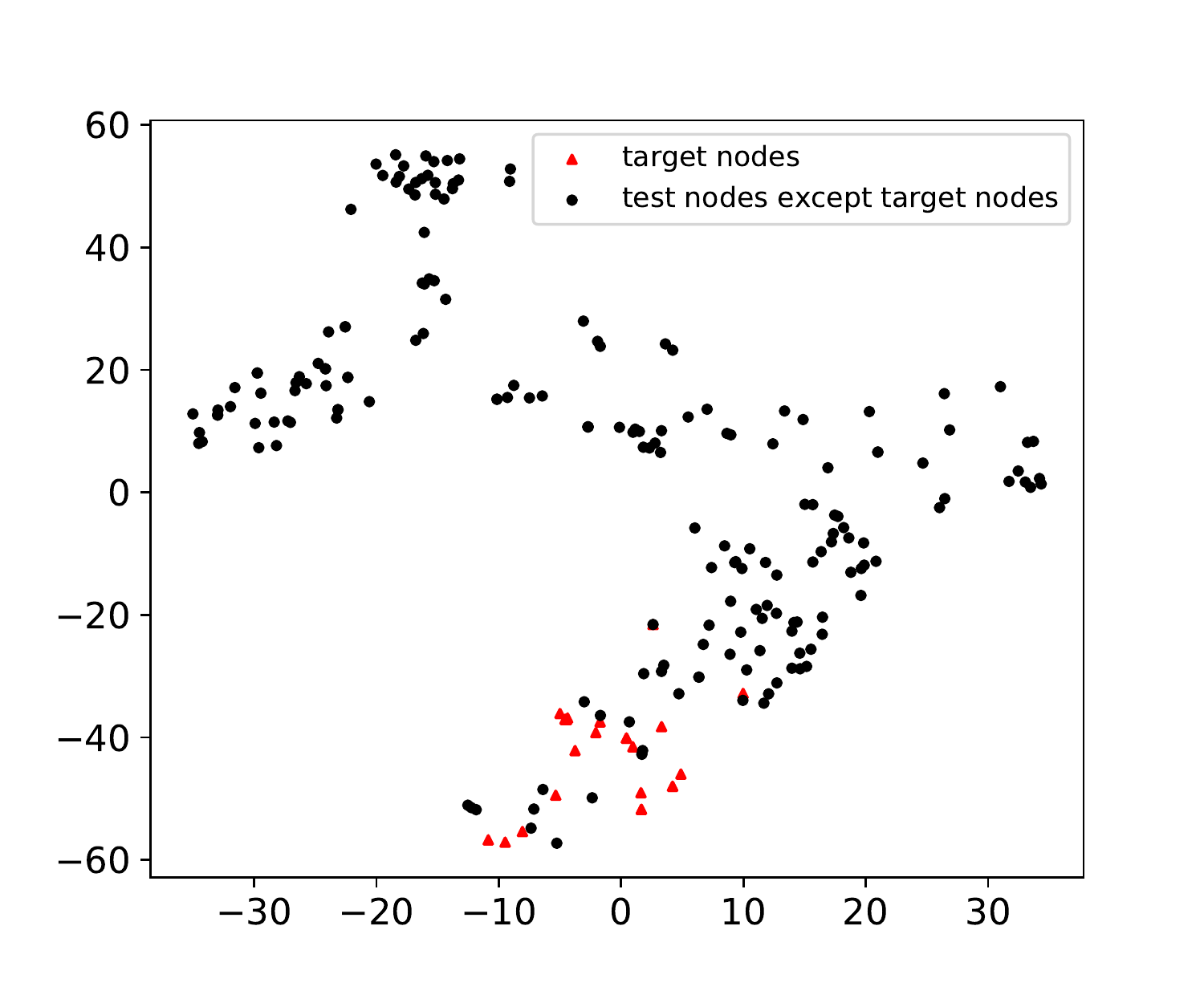}
		\caption{\textbf{Bitcoin-Alpha} with $B=50$}
	\end{subfigure}
	\hfill
	\begin{subfigure}[b]{0.234\textwidth}
		\centering
		\includegraphics[width=\textwidth,height=3cm]{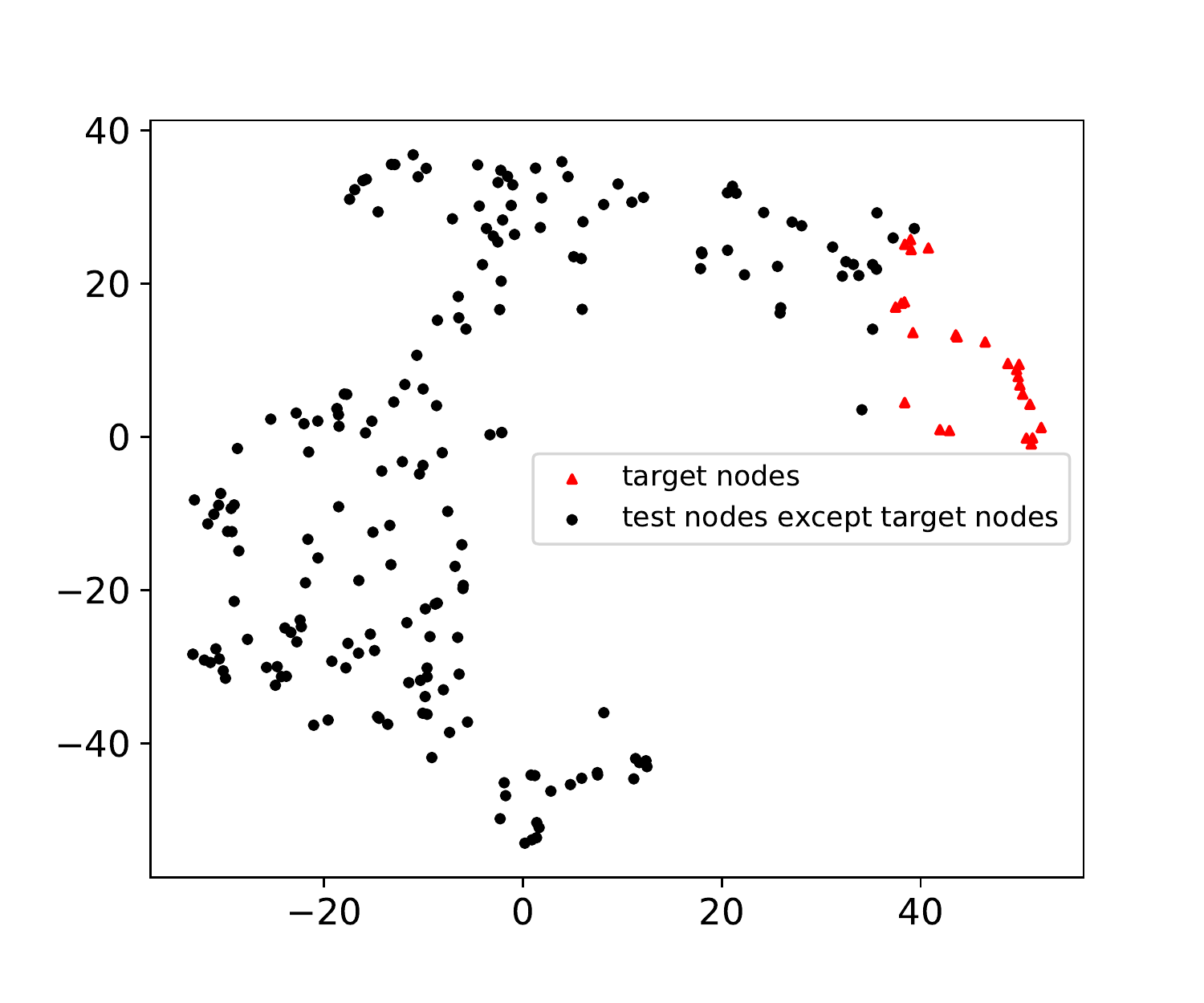}
		\caption{\textbf{Wikivote} clean graph}
	\end{subfigure}
	\hfill
	\begin{subfigure}[b]{0.234\textwidth}
		\centering
		\includegraphics[width=\textwidth,height=3cm]{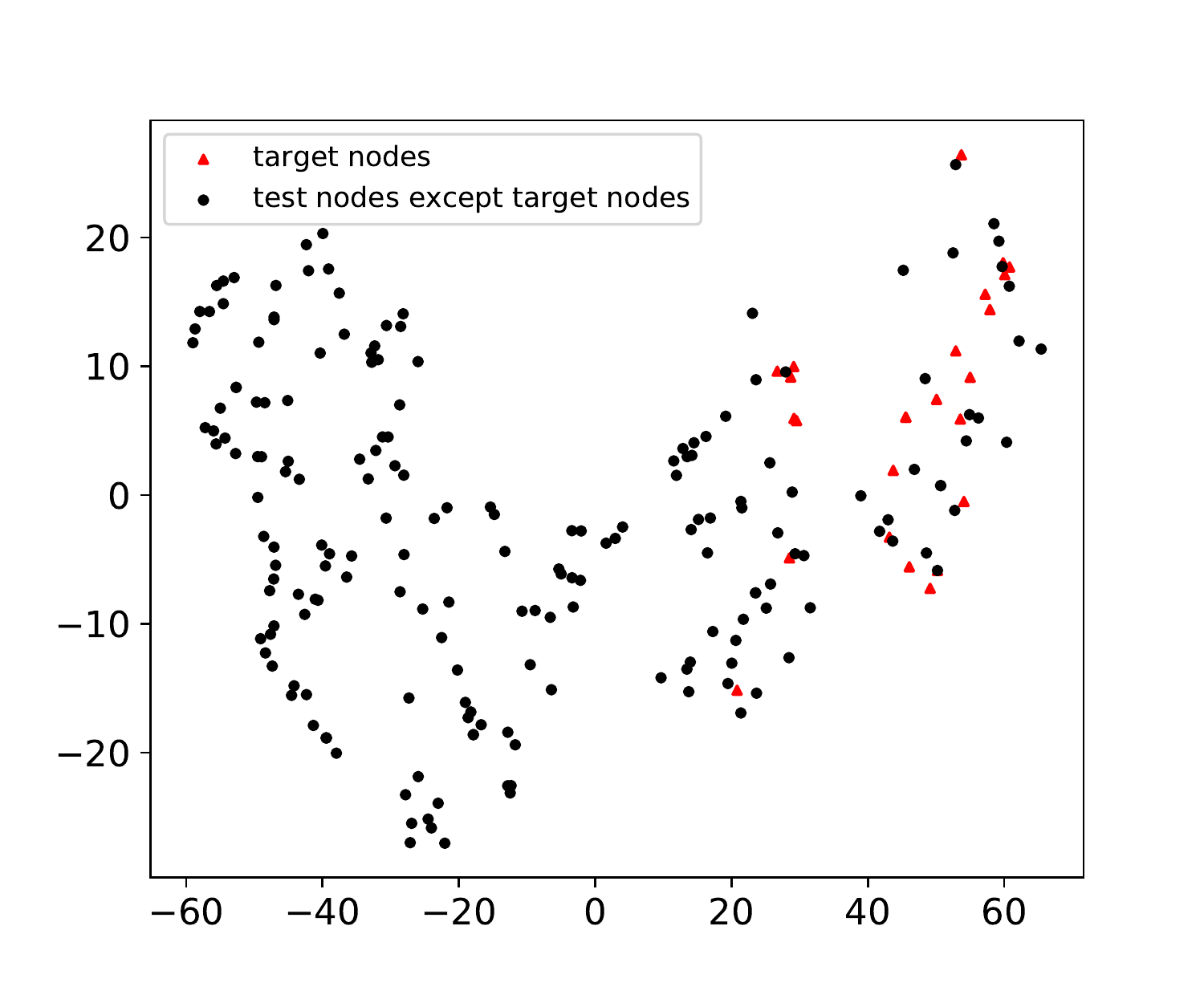}
		\caption{\textbf{Wikivote} with $B=100$}
	\end{subfigure}
	\caption{Scatterplot of ReFeX embeddings of penultimate hidden features for clean graph and poisoned graph of \textbf{Bitcoin-Alpha} and \textbf{Wikivote} in 2-D space}
	\label{fig-emb-refex}
\end{figure}

Ideally, we can expect that before attack, there is a clear linear decision boundary between the anomalous nodes and the benign nodes. After the attack, the anomalous nodes are mixed in the benign ones. Fig.~\ref{fig-emb-gal} and Fig.~\ref{fig-emb-refex} shows that $\mathsf{BinarizedAttack}$ can indirectly influence the quality of the node embeddings and thus break the linear separable decision boundary of the anomaly detection system.

\subsection{Countermeasures against $\mathsf{BinarizedAttack}$}
\begin{figure}[h]
	\centering
	\begin{subfigure}[b]{0.234\textwidth}
		\centering
		\includegraphics[width=\textwidth,height=3cm]{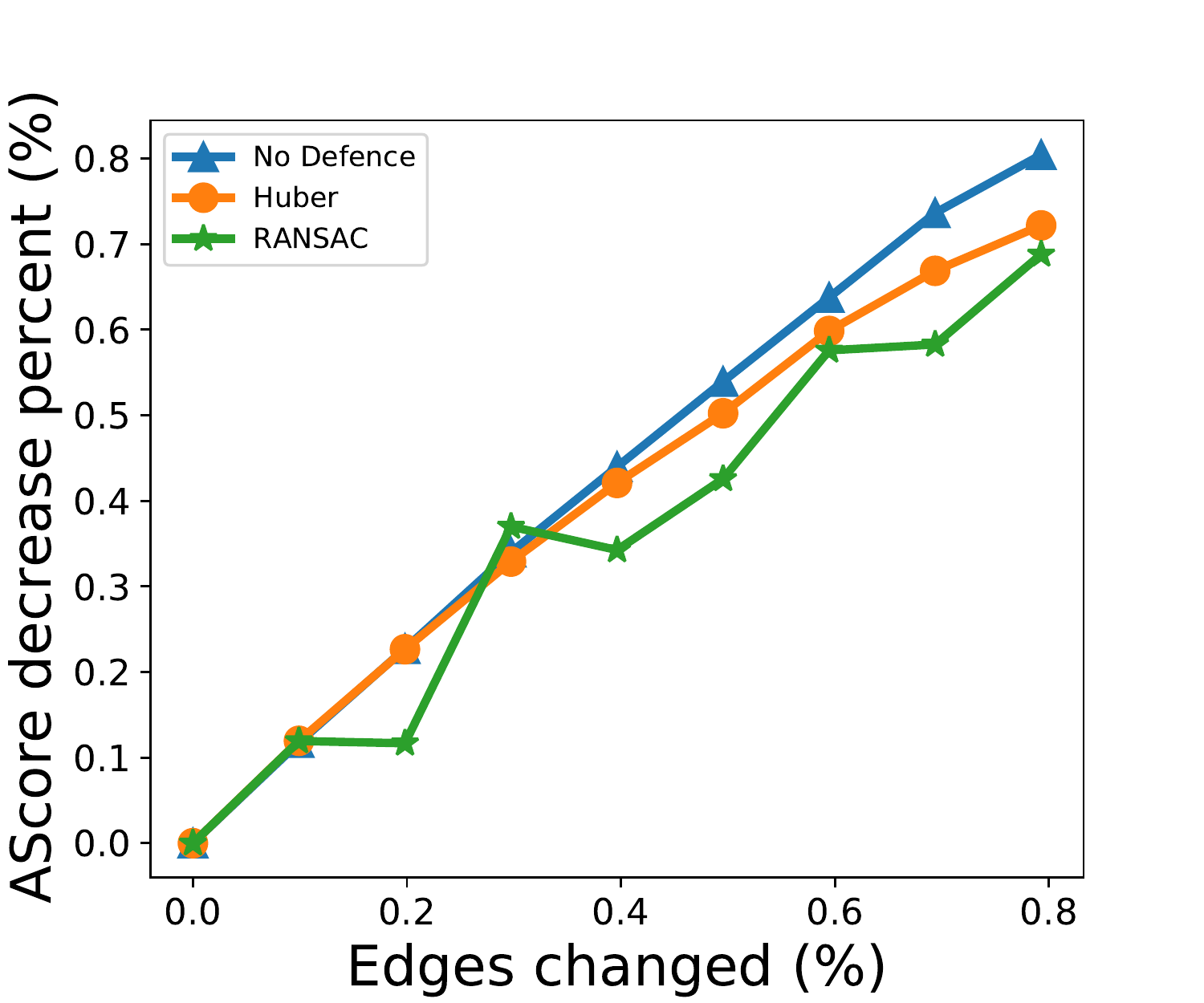}
		\caption{\textbf{Bitcoin-Alpha}}
	\end{subfigure}
	\hfill
	\begin{subfigure}[b]{0.234\textwidth}
		\centering
		\includegraphics[width=\textwidth,height=3cm]{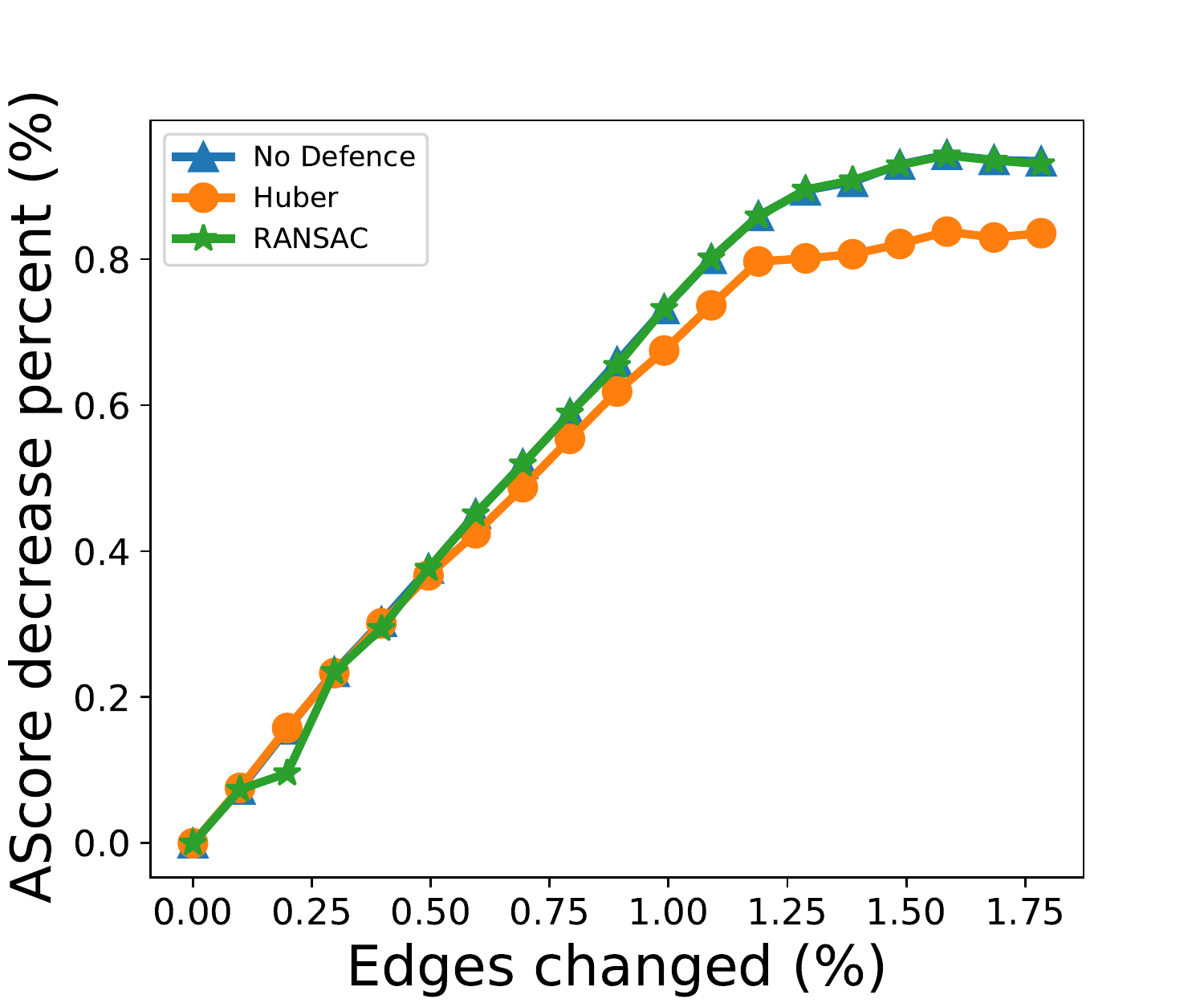}
		\caption{\textbf{Wikivote}}
	\end{subfigure}
	\caption{Defence against $\mathsf{BinarizedAttack}$}
	\label{fig-defense}
\end{figure}

In this part, we further analyze the possible defence method against $\mathsf{BinarizedAttack}$. We retrain the linear model based on the clean graph and poisoned graphs with different perturbations by Huber regressor \cite{10.1214/aoms/1177703732} and RANSAC \cite{ransac} method. We compare the $\mathsf{AScores}$ of two robust estimations with OLS estimation. The experiments are implemented based on the two real datasets: \textbf{Bitcoin-Alpha} and \textbf{Wikivote}. The attack method is $\mathsf{BinarizedAttack}$. We randomly select 10 nodes as our target nodes for $\mathsf{BinarizedAttack}$.  For each of the dataset, we implement the experiments $5$ times respectively and report the percentage of the decreasing of the total $\mathsf{AScores}$ for the target set. The experimental results are shown in Fig.~\ref{fig-defense}. 

From Fig.~\ref{fig-defense} we observe that both Huber regression and RANSAC can slightly mitigate the attack. But overall, $\mathsf{BinarizedAttack}$ remains a very effective attack. We leave the defense of $\mathsf{BinarizedAttack}$ as future work.

\section{Conclusion}
\label{sec-conclude}
In this paper, we initiate the study to investigate the vulnerability of graph-based anomaly detection under structural poisoning attacks. Specifically, we use $\mathsf{OddBall}$ as a representative target GAD system and mathematically formulate the attacks against it as a bi-level optimization problem. To address the technical challenge of solving the combinatorial optimization problem, we propose a novel gradient-descent-based method $\mathsf{BinarizedAttack}$, which effectively enables target nodes to evade anomaly detection, significantly outperforming existing methods. We further use $\mathsf{BinarizedAttack}$ to attack other representation-learning-based GAD systems in a black-box setting. Results show that $\mathsf{BinarizedAttack}$ can also evade those systems that it is not designed for, greatly relaxing the information requirement for the attackers. Overall, our study demonstrates that current GAD systems are indeed vulnerable to structural poisoning attacks. Our research also opens the door to investigating the adversarial robustness of other graph-based security analytic tools under this new type of attacks. We leave the defense of structural attacks as an important direction of future works.



\bibliographystyle{ieeetr}
\bibliography{citation}

\begin{thebibliography}{10}

\bibitem{8621913}
S.~Khaled, N.~El-Tazi, and H.~M.~O. Mokhtar, ``Detecting fake accounts on
  social media,'' in {\em 2018 IEEE International Conference on Big Data (Big
  Data)}, pp.~3672--3681, 2018.

\bibitem{fraudpayments}
A.~Diadiushkin, K.~Sandkuhl, and A.~Maiatin, ``Fraud detection in payments
  transactions: Overview of existing approaches and usage for instant
  payments,'' {\em Complex Systems Informatics and Modeling Quarterly},
  pp.~72--88, 10 2019.

\bibitem{180611}
Q.~Cao, M.~Sirivianos, X.~Yang, and T.~Pregueiro, ``Aiding the detection of
  fake accounts in large scale social online services,'' in {\em 9th {USENIX}
  Symposium on Networked Systems Design and Implementation ({NSDI} 12)}, (San
  Jose, CA), pp.~197--210, {USENIX} Association, Apr. 2012.

\bibitem{7791883}
M.~{Iwamoto}, S.~{Oshima}, and T.~{Nakashima}, ``A study of malicious pdf
  detection technique,'' in {\em 2016 10th International Conference on Complex,
  Intelligent, and Software Intensive Systems (CISIS)}, pp.~197--203, 2016.

\bibitem{oddball}
L.~Akoglu, M.~McGlohon, and C.~Faloutsos, ``oddball: Spotting anomalies in
  weighted graphs,'' in {\em Advances in Knowledge Discovery and Data Mining}
  (M.~J. Zaki, J.~X. Yu, B.~Ravindran, and V.~Pudi, eds.), (Berlin,
  Heidelberg), pp.~410--421, Springer Berlin Heidelberg, 2010.

\bibitem{Z_gner_2018}
D.~Zügner, A.~Akbarnejad, and S.~Günnemann, ``Adversarial attacks on neural
  networks for graph data,'' {\em SIGKDD}, Jul 2018.

\bibitem{DBLP:journals/corr/abs-1902-08412}
D.~Zügner and S.~Günnemann, ``Adversarial attacks on graph neural networks
  via meta learning,'' in {\em International Conference on Learning
  Representations}, 2019.

\bibitem{courbariaux2016binarized}
I.~Hubara, M.~Courbariaux, D.~Soudry, R.~El-Yaniv, and Y.~Bengio, ``Binarized
  neural networks,'' in {\em Proceedings of the 30th International Conference
  on Neural Information Processing Systems}, pp.~4114--4122, 2016.

\bibitem{10.1145/3340531.3411979}
T.~Zhao, C.~Deng, K.~Yu, T.~Jiang, D.~Wang, and M.~Jiang, {\em Error-Bounded
  Graph Anomaly Loss for GNNs}, p.~1873–1882.
\newblock New York, NY, USA: Association for Computing Machinery, 2020.

\bibitem{ReFeX}
K.~Henderson, B.~Gallagher, L.~Li, L.~Akoglu, T.~Eliassi-Rad, H.~Tong, and
  C.~Faloutsos, ``It's who you know: Graph mining using recursive structural
  features,'' in {\em Proceedings of the 17th ACM SIGKDD International
  Conference on Knowledge Discovery and Data Mining}, KDD '11, (New York, NY,
  USA), p.~663–671, Association for Computing Machinery, 2011.

\bibitem{akoglu2015graph}
L.~Akoglu, H.~Tong, and D.~Koutra, ``Graph based anomaly detection and
  description: a survey,'' {\em Data mining and knowledge discovery}, vol.~29,
  no.~3, pp.~626--688, 2015.

\bibitem{gao2006converting}
J.~Gao and P.-N. Tan, ``Converting output scores from outlier detection
  algorithms into probability estimates,'' in {\em Sixth International
  Conference on Data Mining (ICDM'06)}, pp.~212--221, IEEE, 2006.

\bibitem{chen2013ascos}
H.-H. Chen and C.~L. Giles, ``Ascos: an asymmetric network structure context
  similarity measure,'' in {\em 2013 IEEE/ACM International Conference on
  Advances in Social Networks Analysis and Mining (ASONAM 2013)}, pp.~442--449,
  IEEE, 2013.

\bibitem{kuang2012symmetric}
D.~Kuang, C.~Ding, and H.~Park, ``Symmetric nonnegative matrix factorization
  for graph clustering,'' in {\em Proceedings of the 2012 SIAM international
  conference on data mining}, pp.~106--117, SIAM, 2012.

\bibitem{kang2011mining}
U.~Kang, D.~H. Chau, and C.~Faloutsos, ``Mining large graphs: Algorithms,
  inference, and discoveries,'' in {\em 2011 IEEE 27th International Conference
  on Data Engineering}, pp.~243--254, IEEE, 2011.

\bibitem{koutra2011unifying}
D.~Koutra, T.-Y. Ke, U.~Kang, D.~H.~P. Chau, H.-K.~K. Pao, and C.~Faloutsos,
  ``Unifying guilt-by-association approaches: Theorems and fast algorithms,''
  in {\em Joint European Conference on Machine Learning and Knowledge Discovery
  in Databases}, pp.~245--260, Springer, 2011.

\bibitem{zhou2020robust}
K.~Zhou and Y.~Vorobeychik, ``Robust collective classification against
  structural attacks,'' in {\em Conference on Uncertainty in Artificial
  Intelligence}, pp.~250--259, PMLR, 2020.

\bibitem{zhou2019attacking}
K.~Zhou, T.~P. Michalak, M.~Waniek, T.~Rahwan, and Y.~Vorobeychik, ``Attacking
  similarity-based link prediction in social networks,'' in {\em Proceedings of
  the 18th International Conference on Autonomous Agents and MultiAgent
  Systems}, pp.~305--313, 2019.

\bibitem{waniek2019hide}
M.~Waniek, K.~Zhou, Y.~Vorobeychik, E.~Moro, T.~P. Michalak, and T.~Rahwan,
  ``How to hide one’s relationships from link prediction algorithms,'' {\em
  Scientific reports}, vol.~9, no.~1, pp.~1--10, 2019.

\bibitem{waniek2018hiding}
M.~Waniek, T.~P. Michalak, M.~J. Wooldridge, and T.~Rahwan, ``Hiding
  individuals and communities in a social network,'' {\em Nature Human
  Behaviour}, vol.~2, no.~2, pp.~139--147, 2018.

\bibitem{jin2020graph}
W.~Jin, Y.~Ma, X.~Liu, X.~Tang, S.~Wang, and J.~Tang, ``Graph structure
  learning for robust graph neural networks,'' in {\em Proceedings of the 26th
  ACM SIGKDD International Conference on Knowledge Discovery \& Data Mining},
  pp.~66--74, 2020.

\bibitem{Z_gner_2019}
D.~Zügner and S.~Günnemann, ``Certifiable robustness and robust training for
  graph convolutional networks,'' {\em Proceedings of the 25th ACM SIGKDD
  International Conference on Knowledge Discovery and Data Mining}, Jul 2019.

\bibitem{wu2019adversarial}
H.~Wu, C.~Wang, Y.~Tyshetskiy, A.~Docherty, K.~Lu, and L.~Zhu, ``Adversarial
  examples for graph data: Deep insights into attack and defense,'' in {\em
  Proceedings of the Twenty-Eighth International Joint Conference on Artificial
  Intelligence, {IJCAI-19}}, pp.~4816--4823, International Joint Conferences on
  Artificial Intelligence Organization, 7 2019.

\bibitem{10.1145/3336191.3371789}
N.~Entezari, S.~A. Al-Sayouri, A.~Darvishzadeh, and E.~E. Papalexakis, ``All
  you need is low (rank): Defending against adversarial attacks on graphs,'' in
  {\em Proceedings of the 13th International Conference on Web Search and Data
  Mining}, WSDM '20, (New York, NY, USA), p.~169–177, Association for
  Computing Machinery, 2020.

\bibitem{Zdaniuk2014}
B.~Zdaniuk, {\em Ordinary Least-Squares (OLS) Model}, pp.~4515--4517.
\newblock Dordrecht: Springer Netherlands, 2014.

\bibitem{Tibshirani94regressionshrinkage}
R.~Tibshirani, ``Regression shrinkage and selection via the lasso,'' {\em
  JOURNAL OF THE ROYAL STATISTICAL SOCIETY, SERIES B}, vol.~58, pp.~267--288,
  1994.

\bibitem{Hoerl1}
A.~E. Hoerl and R.~W. Kennard, ``Ridge regression: Biased estimation for
  nonorthogonal problems,'' {\em Technometrics}, vol.~12, pp.~55--67, 1970.

\bibitem{hamilton2020graph}
W.~L. Hamilton, ``Graph representation learning,'' {\em Synthesis Lectures on
  Artifical Intelligence and Machine Learning}, vol.~14, no.~3, pp.~1--159,
  2020.

\bibitem{kipf2017semisupervised}
T.~N. Kipf and M.~Welling, ``Semi-supervised classification with graph
  convolutional networks,'' in {\em 5th International Conference on Learning
  Representations, {ICLR} 2017, Toulon, France, April 24-26, 2017, Conference
  Track Proceedings}, OpenReview.net, 2017.

\bibitem{DBLP:journals/corr/abs-2006-08900}
A.~Zhang and J.~Ma, ``Defensevgae: Defending against adversarial attacks on
  graph data via a variational graph autoencoder,'' {\em CoRR},
  vol.~abs/2006.08900, 2020.

\bibitem{zhu2019robust}
D.~Zhu, Z.~Zhang, P.~Cui, and W.~Zhu, ``Robust graph convolutional networks
  against adversarial attacks,'' in {\em KDD}, ACM, 2019.

\bibitem{DBLP:journals/corr/abs-1908-07558}
X.~Tang, Y.~Li, Y.~Sun, H.~Yao, P.~Mitra, and S.~Wang, ``Transferring
  robustness for graph neural network against poisoning attacks,'' in {\em
  Proceedings of the 13th International Conference on Web Search and Data
  Mining}, WSDM '20, (New York, NY, USA), p.~600–608, Association for
  Computing Machinery, 2020.

\bibitem{DBLP:journals/corr/abs-2006-08149}
X.~Zhang and M.~Zitnik, ``Gnnguard: Defending graph neural networks against
  adversarial attacks,'' in {\em NeurIPS}, 2020.

\bibitem{10.1214/aoms/1177703732}
P.~J. Huber, ``{Robust Estimation of a Location Parameter},'' {\em The Annals
  of Mathematical Statistics}, vol.~35, no.~1, pp.~73 -- 101, 1964.

\bibitem{ransac}
M.~A. Fischler and R.~C. Bolles, ``Random sample consensus: A paradigm for
  model fitting with applications to image analysis and automated
  cartography,'' {\em Commun. ACM}, vol.~24, p.~381–395, June 1981.

\bibitem{erdHos1960evolution}
P.~Erd{\H{o}}s and A.~R{\'e}nyi, ``On the evolution of random graphs,'' {\em
  Publ. Math. Inst. Hung. Acad. Sci}, vol.~5, no.~1, pp.~17--60, 1960.

\bibitem{Albert_2002}
R.~Albert and A.-L. Barabási, ``Statistical mechanics of complex networks,''
  {\em Reviews of Modern Physics}, vol.~74, p.~47–97, Jan 2002.

\bibitem{zafarani2014users}
R.~Zafarani and H.~Liu, ``Users joining multiple sites: Distributions and
  patterns,'' 2014.

\bibitem{leskovec2010predicting}
J.~Leskovec, D.~Huttenlocher, and J.~Kleinberg, ``Predicting positive and
  negative links in online social networks,'' in {\em Proceedings of the 19th
  international conference on World wide web}, pp.~641--650, 2010.

\bibitem{kumar2018rev2}
S.~Kumar, B.~Hooi, D.~Makhija, M.~Kumar, C.~Faloutsos, and V.~Subrahmanian,
  ``Rev2: Fraudulent user prediction in rating platforms,'' in {\em Proceedings
  of the Eleventh ACM International Conference on Web Search and Data Mining},
  pp.~333--341, ACM, 2018.

\bibitem{permutationtest}
L.~Stanberry, {\em Permutation Test}, pp.~1678--1678.
\newblock New York, NY: Springer New York, 2013.

\bibitem{softlabel}
G.~Hinton, O.~Vinyals, and J.~Dean, ``Distilling the knowledge in a neural
  network,'' in {\em NIPS Deep Learning and Representation Learning Workshop},
  2015.

\bibitem{tsne}
L.~van~der Maaten and G.~Hinton, ``Viualizing data using t-sne,'' {\em Journal
  of Machine Learning Research}, vol.~9, pp.~2579--2605, 11 2008.

\bibitem{DBLP:journals/corr/abs-1808-05385}
Y.~Li, L.~Ding, and X.~Gao, ``On the decision boundary of deep neural
  networks,'' {\em CoRR}, vol.~abs/1808.05385, 2018.

\bibitem{pca}
K.~P. F.R.S., ``Liii. on lines and planes of closest fit to systems of points
  in space,'' {\em The London, Edinburgh, and Dublin Philosophical Magazine and
  Journal of Science}, vol.~2, no.~11, pp.~559--572, 1901.

\end{thebibliography}

\end{document}